\documentclass[10pt]{article}

\usepackage{headerstuff}
\usepackage{wileyheaderstuff}

\newcommand{\thepapertitle}{ExplainFix: Explainable Spatially Fixed Deep Networks}

\begin{document}
\author[123*]{Alex Gaudio\orcidlink{0000-0003-1380-6620}}%
\author[1]{Christos Faloutsos\orcidlink{0000-0003-2996-9790}}%
\author[1]{Asim Smailagic\orcidlink{0000-0001-8524-997X}}%
\author[23]{Pedro Costa\orcidlink{0000-0002-9528-1292}}%
\author[23]{Aur\'elio Campilho\orcidlink{0000-0002-5317-6275}}%

\affil[1]{Carnegie Mellon University, Pittsburgh, PA, USA}%
\affil[2]{Faculdade de Engenharia da Universidade do Porto, Portugal}%
\affil[3]{INESC TEC, Portugal}

\vspace{-1em}
  \date{}

\title{\thepapertitle}
\begingroup
\let\center\flushleft
\let\endcenter\endflushleft
\maketitle
\endgroup


\selectlanguage{english}
\begin{abstract}

  Is there an initialization for deep networks that requires no learning?  \ExplainFix adopts two design principles: the \fixedFilters principle that \textit{all} spatial filter weights of convolutional neural networks can be fixed at initialization and never learned, and the \nimbleness principle that only few network parameters suffice.
  We contribute (a) \textit{visual model-based explanations}, (b) \textit{speed and accuracy gains}, and (c) \textit{novel tools} for deep convolutional neural networks.
  \ExplainFix gives key insights that spatially fixed networks should have a steered initialization, that spatial convolution layers tend to prioritize low frequencies, and that most network parameters are not necessary in spatially fixed models.  \ExplainFix models have up to \textit{100x} fewer spatial filter kernels than fully learned models and matching or improved accuracy.
  Our extensive empirical analysis confirms that \ExplainFix guarantees nimbler models (train up to \textit{17\%} faster with channel pruning), matching or improved predictive performance (spanning 13 distinct baseline models, four architectures and two medical image datasets), improved robustness to larger learning rate, and robustness to varying model size.
  We are first to demonstrate that \textit{all} spatial filters in state-of-the-art convolutional deep networks can be fixed at initialization, not learned.
\end{abstract}

\sloppy


\newcommand{\FigExplainSteer}{
\begin{tikzpicture}
    \draw (0, 0) node[inner sep=0] {
      \includegraphics[width=.49\linewidth]{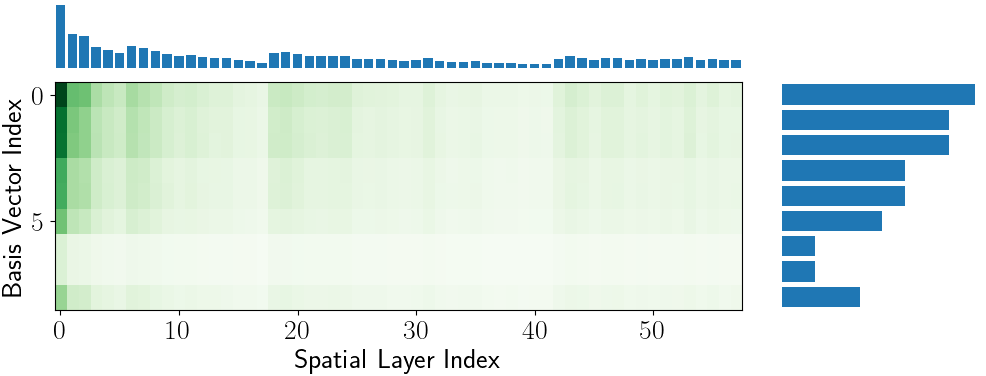}
      };
    \draw [-latex](-3.45,.25) -- (-2.750,-.20);
  \node[rectangle,draw,minimum width=.51cm, minimum height=1.37cm, line width=2, color=purple] (r) at (-3.40,.25) {};
    \draw (-.5, 0) node {\footnotesize 1. Early spatial layers matter most};
    \draw (-.8, -.5) node {\footnotesize 2. Low frequencies dominate};
\end{tikzpicture}
}

\section{Introduction}\label{sec:introduction}

Do all weights of deep convolutional neural network (CNN) need to be learned?  On medical image data, the answer is ``no.''
We analyze and explain the spatial convolution layers internal to CNNs in order to understand how to initialize and prune spatially fixed networks.
\ExplainFix adopts two design principles for fixed weight networks.
The first proposed principle, \fixedFilters, dictates
that we do not learn the spatial convolution filters in the CNN.
The second principle,
\nimbleness, dictates that only a small
fraction of spatial filters suffices.
\ExplainFix has the following main contributions:
\begin{enumerate}
  \item {\bf Model-based Explanations:} \ExplainFix visualizes a model's internal properties, namely spatial filter steering and weight saliency, to show how to improve models according to our design principles.
  \item {\bf Speed and Accuracy Gains}: \ExplainFix models are faster and smaller, with matching or improved accuracy, thanks to our initialization and pruning methods for spatially fixed networks.
  \item {\bf Novel Tools}: The \ExplainFix system comprises open source tools for fixed filter networks, including ExplainSteer, ChannelPrune, and fixed kernel initialization methods (GuidedSteer, GHaar, Psine, Unchanged, DCT2).  
\end{enumerate}

\begin{figure*}[t]
  \centering
  \begin{minipage}{\textwidth}
    \centering
    \subfloat[Visual Model-based Explanations\label{fig:explainsteer_fig1}]{
      \FigExplainSteer
    }\qquad
    \subfloat[Few Spatial Filters Suffice\label{fig:unnecessary_inference_fig1}]{
    \includegraphics[width=.40\textwidth]{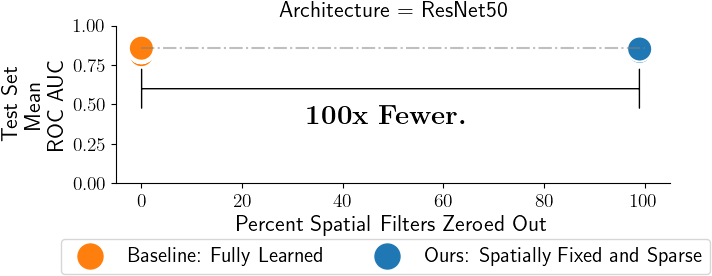}
  }
  \end{minipage}
  \begin{minipage}{.5\textwidth}
    \vspace{.4cm}
    \centering
    \subfloat[Smaller and Faster To Train\label{fig:compute_efficiency}]{
      \includegraphics[width=.95\linewidth]{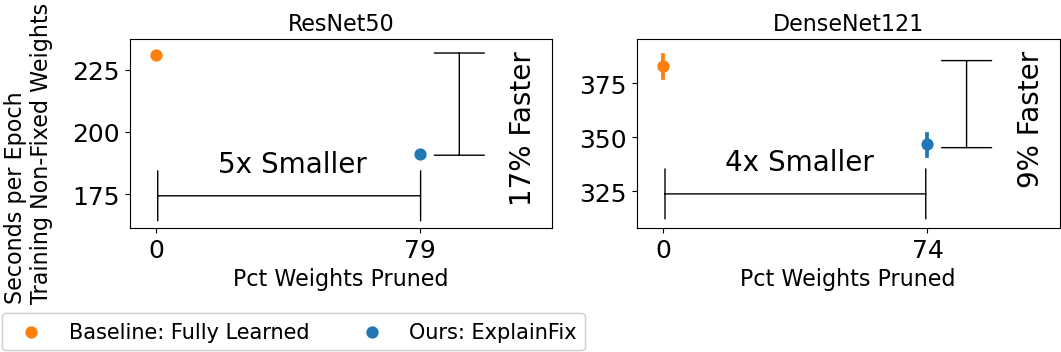}
    }
  \end{minipage}%
  \begin{minipage}{.5\textwidth}
    \vspace{.4cm}
    \centering
    \subfloat[Matching Accuracy\label{fig:good_accuracy}]{
      \includegraphics[width=.95\linewidth]{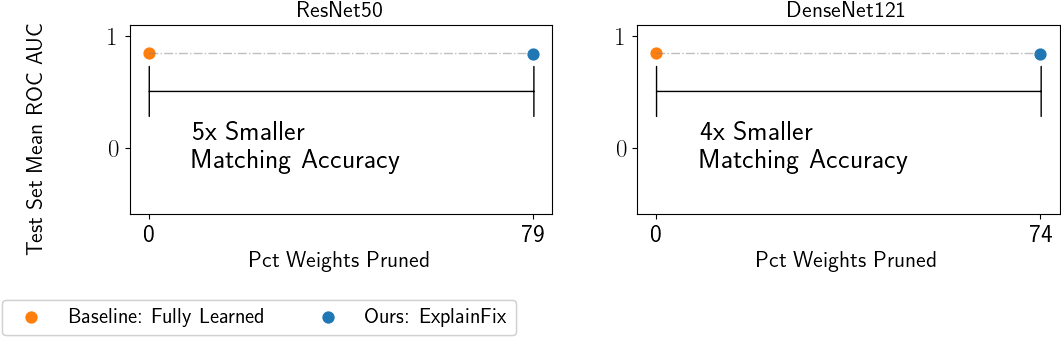}
    }
  \end{minipage}
  \captionof{figure}[c]{
    \textbf{\ExplainFix checks all goals: explainable (a,b), nimble (c), and accurate (d).}
    \textbf{(a)} Our ExplainSteer visual explanation exposes deep network inefficiencies.
    \textbf{(b)} Only 1 in 100 filters needs to be kept on ResNet50.  We have similar results on other architectures.
    \textbf{(c,d)} Our ChannelPrune method makes models smaller and faster (c) and accurate (d).
  }\label{fig:bestplot}
\end{figure*}

  Fig. \ref{fig:bestplot} demonstrates the value of our \ExplainFix system.  
  Fig.  \ref{fig:explainsteer_fig1} presents our visual explanation of a CNN to highlight that nearly all spatial layers can be pruned.
  Fig. \ref{fig:unnecessary_inference_fig1} validates the visual explanation to show that we can use 100x fewer spatial filters while preserving predictive performance.  We demonstrate in Sec. \ref{sec:results_pruning} that most spatial filters are indeed unnecessary for both training and inference.
  Fig. \ref{fig:compute_efficiency} illustrates that a spatially fixed architecture pruned by our method is faster and smaller than the baseline, and Fig. \ref{fig:good_accuracy} shows the model is nearly equal in accuracy.  Similar results hold for other models in Sec. \ref{sec:compsave}.

Our hypothesis and conclusion that \textit{all} spatial 2D convolution kernels can be fixed with a steered initialization, not learned, and mostly eliminated is strongly supported by our insights, explanations and results.  In Sec. \ref{intuitions}, we review related work and provide intuitions about relationships between deep networks and steerability.
We describe our initialization, explanation and pruning methods in Sec. \ref{methods}, and the experiments and resulting analyses in Sec. \ref{results}.  In Sec. \ref{discussion}, we discuss applications and possible future research directions for fixed filter networks.  To our knowledge, we are first to consider entirely fixed spatial convolution filters and their initialization methods without changing deep network architecture.

\section{Background and Related Work}\label{intuitions}

The proposed \ExplainFix is the only method with all properties in Table \ref{tab:salesman}.
We highlight current challenges in deep learning and benefits of a fixed weight initialization.  First, deep networks are increasingly larger \parencite{deeplearninglargerisbetter}, they require large datasets that are costly to acquire and annotate \parencite{annotationsexpensive,bioimageannotationsdifficulttoanalyzeandget}, and the models are difficult to train.  The unsustainable increasing trends call for dramatically more efficient methods \parencite{deeplearningcomputationallimits}.  A fixed weight initialization minimizes the need for training, computational resources and data.

A second major challenge in deep learning is lack of interpretability.  Despite gains, deep networks are still considered black boxes \parencite{blackbox_survey_xai,survey_medical_xai,pleasemakedeepnetsmoreexplainable}.  
Ante-hoc methods create novel deep network architectures that are explainable by their design \parencite{capsulenetworks_survey2019,a_good_blackbox_survey_xai_on_arxiv,bruna2013scatteringthesis,pixelcoloramplification,angelov2020towards}, yet these methods require up-front knowledge of how to design the network.  We propose to develop a model-based explanation of spatial convolution layers that is compatible with many ante-hoc methods.  Ante-hoc explainable fixed filter methods have been rigorously proposed, for example, in wavelet scattering networks \parencite{mallatscatteringformal,cotter2019scatternet,oyallon2018scattering,bruna2013scatteringthesis} and in group equivariant convolution networks \parencite{cohen2016group_equivariant_conv}.  They yield interpretable, state-of-the-art networks and can help to understand how deep networks learn \parencite{mallat2016understanding,cotter2017visualizing}, but these models are hard to use.  The co/in/equi-variance properties must be fully specified in advance, and they also introduce custom network architecture, thus discouraging wider adoption.  We therefore contribute methods that start with a standard convolutional network architecture and do not need advance knowledge of the co/in/equi-variance properties or the dataset.
Finally, a recent empirical study found that fixing as many as 90\% of weights across the network, without discerning between types of convolutions or layers, had slightly decreased performance on common datasets \parencite{rosenfeld2019_fixedfilters}.  Despite showing reduced performance, the results justify our work.  We propose to fix all spatial convolution layers and we develop explanations and methods for how to initialize the weights and prune channels without performance loss.

\textbf{Redundancy: Some weights are unnecessary.}
Neural networks are highly compressible \parencite{compressionsurvey}.  Pruning 99\% of nodes on a trained multi-class model can yield no loss in single-class performance \parencite{leino2018influence}. Pruning the connections between nodes by 9-13x \parencite{deepcompression} is possible without performance loss.  Temporarily removing nodes during training with Dropout can improve performance and prevent over-fitting \parencite{dropout}.  The recently popular Lottery Ticket Hypothesis states that over-parameterized and untrained neural networks contain subnetworks, or winning tickets, that enjoy roughly equal accuracy to the original trained baseline network, either with training \parencite{lottery_ticket_hypothesis_frankle2018} or without any training \parencite{lottery_ticket_notraining_original_proof_malach2020,lottery_ticket_notraining_better_proof_pensia2021}.
Moreover, over-parameterized deep networks contain a frozen subspace of weights that are not changed by learning with gradient descent \parencite{saxe_generalization2020}, and fixing as much as 90\% of randomly sampled weights in a deep network was shown to give small performance loss \parencite{rosenfeld2019_fixedfilters}.  Thus, deep networks are highly redundant, with a subset of weights relevant to the predictive task, and a subset of weights unchanged by learning.  In this paper, we offer empirical evidence that \textit{all spatial filters} can be entirely unchanged by learning and that as few as only 1\% of spatial filters are necessary for the inference task.

\textbf{Steerability: A property of convolution layers.}  
A steered filter is a linear combination of basis vectors \parencite{steerablefilters}.
The convolution layer in deep networks, as in Eq. \ref{eq:steerability}, defines each output channel as a sum of linearly transformed input channels, where $\bO_o$ is an output channel, $\I_i$ is an input channel, $\f_{o,i}$ are the weights of a convolution filter kernel, and $*$ is the cross-correlation operator.  By a change of variables inside the sum, each output channel $\bO_o$ can be seen as a steered combination of either the normalized filters $\tilde\f_{o,i}$, the feature maps $\tilde\f_{o,i} * \I_i$, or the input channels $\I_i$, via the steering weights $w_{o,i}=\norm{\f_{o,i}}$.  Thus, convolution layers have a steered representation by design.  Redundancy by steering is also an essential property of wavelet scattering networks \parencite{bruna2013scatteringthesis}.  Our work provides empirical evidence that steered representation is an important property of spatially fixed CNNs.
\begin{align}
  \bO_o &= \sum_i \f_{o,i} * \I_i 
  = \sum_i \left(\norm{\f_{o,i}} \frac{\f_{o,i}}{\norm{\f_{o,i}}}\right) * \I_i
        = \sum_i w_{o,i} (\tilde\f_{o,i} * \I_i) = \sum_i \tilde\f_{o,i} * (w_{o,i}\I_i) \label{eq:steerability}
\end{align}

\textbf{The pointwise $1\times1$ convolution} was introduced and popularized in \parencite{network_in_network,inceptionv3,resnet,mobilenetv1}.  A $1\times1$ convolution reduces the filter kernel in Eq. \ref{eq:steerability} to the scalar $\tilde f=1$.  Each output channel $\bO_o$ is a linear mixture of the input channels, via mixing weights $w_{o,i}$.  We do not consider the $1\times1$ convolution a spatial convolution.  Eq. \ref{eq:steerability} implies that $1\times1$ convolutions can also steer the inputs or outputs of a nearby spatial convolution.  Thus, both network architectures and spatial convolution layers can jointly learn an optimal steering of spatial features.  Our results suggest that relying on architecture alone is insufficient.  Initializing fixed spatial convolution layers with redundant, steered fixed initializations can facilitate training of non-fixed weights and give better prediction performance.

\textbf{Differences between Deep Networks and Wavelets:}
A discrete wavelet transform (DWT) \parencite{daubechies1992tenlectures} can be described as a sequence of convolution operations with fixed wavelet filters that map a single input channel to multiple output channels. The convolution layer in deep networks has minor differences; it uses a cross-correlation operator (filters are flipped) rather than a convolution operator, the filters are learned, and each output channel is a sum of the linearly transformed input channels.  The sum over multiple input channels is entirely captured by steerability in Eq. \ref{eq:steerability}.  The other difference, that filters are learned, is the main subject of our hypothesis.  

\newcommand{\goodcheck}{\CheckmarkBold}
\newcommand{\badcheck}{}
\newcommand{\qmark}{\textbf{?}}
\newcommand{\equals}{\textbf{=}}
\newcommand{\specialcell}[2][c]{\rotatebox{90}{\begin{tabular}[#1]{@{}l@{}}#2\end{tabular}}}

\begin{table}[t]
  \centering
   \caption{\textbf{\ExplainFix Wins.} Fixed Spatial Filters are preferable to existing approaches.}\label{tab:salesman}
\begin{center}
  \small
\begin{tabular}{ l| c |c| c | c || c |}
  Property / Method
  & \specialcell{Wavelet Scattering\\\cite{brunainvariantscatteringnetwork}\\\cite{oyallon2018scattering}\\\cite{cotter2019scatternet}} 
  & \specialcell{Deep Wavelet Networks\\\cite{de2020deep}\\\cite{gabor_conv_nets}\\\cite{gabor_layers_early_help_robustness2020}} 
  & \specialcell{Network Pruning\\\cite{rosenfeld2019_fixedfilters}\\\cite{deepcompression}\\\cite{dropout}\\\cite{lottery_ticket_hypothesis_frankle2018}}
  & \specialcell{Deep Networks \\ \cite{resnet}\\\cite{efficientnet}\\\cite{densenet}}
  & \specialcell{Explainable Fixed\\Filter Networks}
  \\ \hline  
Model Interpretability & \goodcheck & \goodcheck & \badcheck & \badcheck & \goodcheck \\
Faster Training & \goodcheck & \badcheck & \qmark & \equals & \goodcheck \\
Lower RAM Footprint & \goodcheck & \badcheck & \goodcheck & \equals & \goodcheck \\
No Performance Loss & \badcheck & \goodcheck & \qmark & \equals & \goodcheck \\
\hline
Compatible with \ExplainFix & \goodcheck & \goodcheck & \goodcheck & \goodcheck & \goodcheck
\end{tabular}
\end{center}
\end{table}

\begin{figure}[b]
  \centering
    \includegraphics[width=\linewidth]{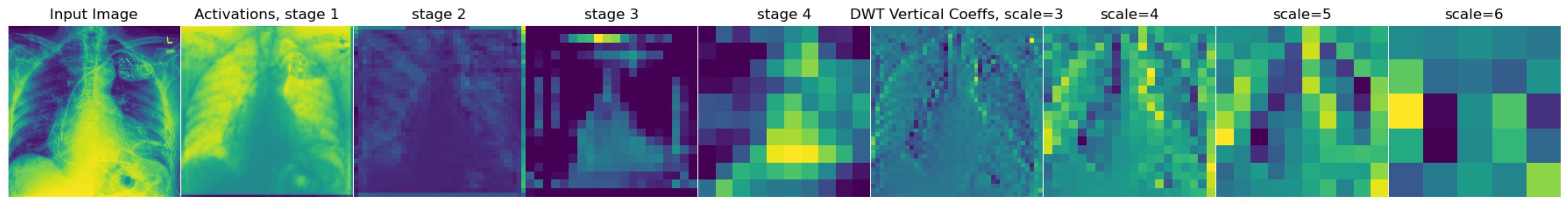}
    \caption{\textbf{Interpreting Deep Networks and Wavelets as edge detectors.} Compare randomly sampled activations from ResNet (left) to a discrete wavelet transform (right).}\label{fig:activations}
\end{figure}

\textbf{Similarities to Discrete Wavelet Transforms:}
Many saliency based explanations of deep networks, or methods that use the deep network's own parameters and activations to explain the network itself, behave analogously to edge detectors \parencite{sanitychecksedgedetector}.  The association of convolutions to edge detection was well known even thirty years ago \parencite{caelli1988filter} for computing wavelet transforms \parencite{waveletsedgedetectors}, and wavelets have excellent edge detection properties \parencite{waveletsedgedetectors2}.  
Deep networks and wavelets both perform edge detection.  Fig. \ref{fig:activations} shows that activations randomly sampled from a pre-trained ResNet50 \parencite{resnet} and vertical coefficients of a Haar DWT both generate edge features.  
Gabor wavelets have been studied in deep networks at all layers \parencite{gabor_conv_nets}, and applied in early layers to enhance a network's robustness to adversarial attack \parencite{gabor_layers_early_help_robustness2020}.  Wavelets have been used to help train autoencoders \parencite{said2016deep} or sparse autoencoders \parencite{mallick2019brain}, where the encoder is subsequently used for classification.  Depthwise separable convolutions \parencite{mobilenetv2} apply a sequence of layers: $1\times 1$ for channel expansion, $3\times 3$ grouped convolution, $1\times 1$.  Analogously to a DWT, the first two convolutions also map each input channel to multiple outputs. The authors of \parencite{mobilenetv2} proposed no ReLU non-linearity after the last convolution to preserve information from negative values, a fact in harmony with the wavelet admissibility constraint that wavelet filters have zero mean (and therefore output negative values).  Additionally, dilated convolution networks \parencite{CANdilatedconv} and the \^{a} trous wavelet transform (AWT) \parencite{atrous_and_mallat_and_discrete_wavelets} both use dilated convolutions, and the former even proposes a nearly fixed initialization.  Atrous Spatial Pyramid Pooling (ASPP) from DeepLabV3 \parencite{deeplabv3} and a multi-scale AWT both apply dilated convolutions in parallel.  ASPP has an analog to a first order wavelet scattering network due to its use of non-linearities.  Wavelet scattering networks have been proposed as an interpretation of the basic structure of a convolutional network \parencite{brunainvariantscatteringnetwork}.  They place a custom structure in early \parencite{oyallon2018scattering} and middle layers \parencite{cotter2019scatternet}, while ASPP is placed at the end.  The deep wavelet neural network of~\parencite{de2020deep} is essentially a one-level wavelet scattering transform with a multi-scale DWT but lacking explainability.  Namely, it recursively applies a set of orthogonal wavelet filters and a downsampling step to an input image; the output is pooled to extract features for use with a classifier like a support vector machine or extreme learning machine.  Moreover, deep convolutional networks have been interpreted as a multilayer implementation of a convolutional wavelet frames \parencite{kang2018deep}. These many similarities suggest deep network spatial filters can be initialized and fixed with wavelet-like filters.

\textbf{Fixed spatial filter networks are preferable to existing related works.}
Table \ref{tab:salesman} highlights four criteria where spatially fixed methods improve deep networks more effectively than the existing approaches.  \ExplainFix is the only approach that satisfies all criteria.  Wavelet Scattering Networks underperform on medium and large datasets and impose specific architectures that significantly limit their incorporation into standard deep networks.
Deep wavelet networks add computational overhead by introducing new architecture, they are often not helpful in later layers, and spatial convolution filters are typically learned.
Network Pruning methods prioritize efficiency over explainability and they do not necessarily have faster training or preserve predictive performance.  Deep Networks are the reference baseline; they are fully learned blackbox models.  We denote an equals sign where the baseline compares to itself by definition.
Our proposed \ExplainFix method satisfies all criteria, and \ExplainFix visual explanations are mutually compatible with all methods.

\section{Proposed Methods}\label{methods}
The proposed methods are ordered into three subsections corresponding to the organization and development of our experiments:  steerability, visual model-based explainability, and pruning.  
Sec. \ref{methods_fixed_filter} introduces methods for showing that steered initialization is important to good performance of spatially fixed networks.  We propose six spatial convolution initialization methods.  
The methods Ones and DCT2 are not steered, while Unchanged, GHaar, Psine, and GuidedSteer are steered.  GHaar, Psine and GuidedSteer are novel contributions, and comprise three distinct ways to steer a network.  GHaar linearly steers a sinusoid basis.  Psine uses a polynomial steering of the basis.  GuidedSteer initializes to the steering properties of a guide model.
In order to visualize and verify our proposed initializations, we develop ExplainSteer in Sec. \ref{methods_explainsteering} as a novel visual model-based explanation tool.  The ExplainSteer visualization shows at a glance the initialization of a network to verify the properties we designed for, and to enable sanity checks that spatial weights are indeed fixed before and after training.  We extend ExplainSteer by incorporating a saliency weight, and we show how the explanation with saliency usefully highlights major inefficiencies in the design of deep network architectures.  Finally, we exploit our saliency weighted visual model-based explanations by developing a saliency-based pruning method ChannelPrune in Sec. \ref{sec:pruning} to systematically remove unnecessary channels across a convolutional network, resulting in a smaller and faster deep network.  The \ExplainFix system comprises three components: fixed initialization methods, model-based visual explanation, and pruning.

  \subsection{Proposed Spatial Filter Initialization for Steerability Testing}\label{methods_fixed_filter}

  \textbf{Ones} sets all spatial filter weights to one.  The ones filter, also called a box filter, approximates a Gaussian blur.  Wavelet transforms use box filters to obtain approximation coefficients.  Ones is not steered.  It tests our assumption that spatial filters need more complex initialization. Ones is visualized in Appendix \ref{appendix_DCT-II} as a matrix with label ``0''.

  \textbf{DCT2} initializes each spatial filter to a randomly selected orthonormal basis filter of the Discrete Cosine Transform Type II (DCT-II), defined in Sec. \ref{methods_explainsteering} and visualized in Appendix \ref{appendix_DCT-II}.  DCT2 is not steered.  It  
  enables testing the hypothesis that fixed spatial filter kernels should be steered.

\textbf{Unchanged} freezes the spatial filters of the model.  We define \textit{Random Unchanged} as spatial filters from a Kaiming Uniform random, never trained network.  \textit{ImageNet Unchanged} fixes spatial filter weights to their pre-trained values.  We analyze how both initializations are steered in Sec. \ref{results_explainability} and Appendix \ref{appendix:saliency_vs_not}.  Visual examples of the Unchanged filter are shown in Appendix \ref{sec:appendix_Unchanged}.

\begin{figure}[t]
  \centering
  \begin{minipage}{.5\textwidth}
  \begin{minipage}[b]{.334\linewidth}
    \footnotesize
    \begin{align*}
      \ba &= \begin{bmatrix} 1 & 1 \end{bmatrix}^\top
        \\
      \bb &= \begin{bmatrix} 1 & -1 \end{bmatrix}^\top
      \\\\
      \ba &= \begin{bmatrix} 1 & 1 & 1 \end{bmatrix}^\top
        \\
      \bb &= \begin{bmatrix} 1 & 0 & -1 \end{bmatrix}^\top
    \end{align*}%
  \end{minipage}%
      \includegraphics[width=.6\linewidth]{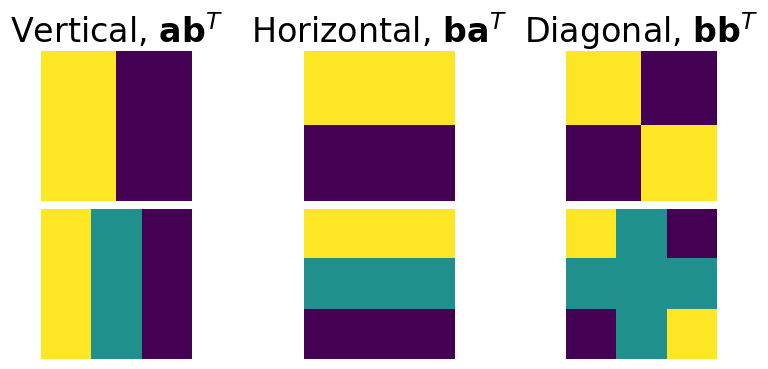}
      \caption{\textbf{Proposed 2D GHaar Basis.}  \\Top row: 2D Haar basis. \\Bottom row: 2D GHaar basis for $3\times3$ filters. \\``Ones'' matrix $\ba\ba^\top$ not shown.}
  \label{fig:haar}
  \end{minipage}
  \begin{minipage}{.49\textwidth}
  \begin{algorithm}[H]
  \caption{Layer-wise GuidedSteer Algorithm.}\label{algo:guidedsteer}
  \small
  \begin{algorithmic}[1]
    \Procedure{GuidedSteer}{$n, \ell, G_1, \dots, G_L$}
  \State Inputs: $n$ desired outputs, $G_{\ell}$ guide filters at layer $l$
    \State $G \gets \texttt{concatRows}(G_1, \dots, G_L)$
    \State $\_, \_, B \gets \texttt{svd}(G^\top G)$
    \For {$\text{col} \in G_\ell B^\top$}
      \State $\w_i^\top \gets \text{KDE(col).resample(n\_times=n)}$
    \EndFor
    \State $W \gets \text{concatColVecs}([\w_1, \dots, \w_m])$
    \State $F_\ell \gets WB + 0$
    \State \textbf{return} $F_\ell \in \mathbb{R}^{n,m}$
    \EndProcedure
  \end{algorithmic}
\end{algorithm}
  \end{minipage}
\end{figure}

\begin{figure}[t]
    \begin{minipage}{.48\linewidth}
        \centering
        \includegraphics[width=\linewidth]{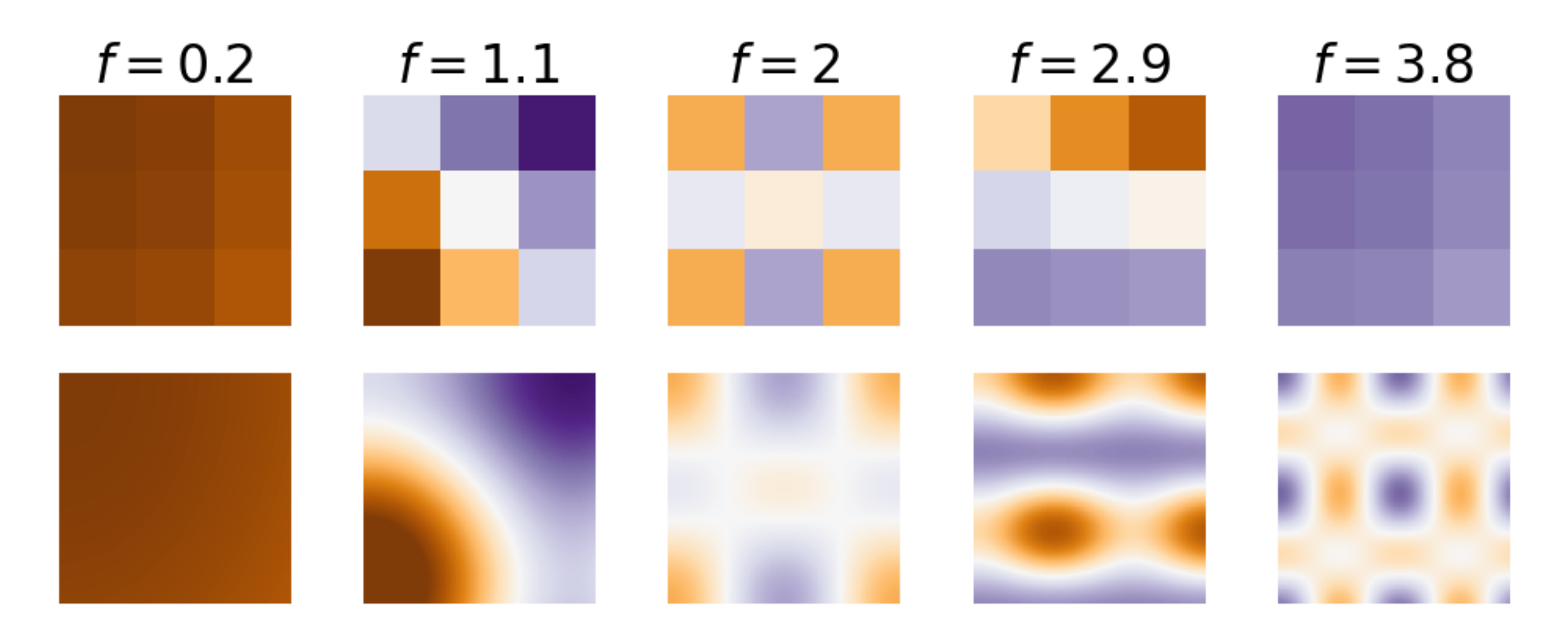}
        \caption{\textbf{GHaar Filters} visualized as 3$\times$3 filters (top row) and in high resolution (bottom row) to give an intuitive sense of the filter.  The frequencies and steering weights are randomly chosen.}\label{fig:ghaar}
    \end{minipage}\hfill
    \begin{minipage}{.48\linewidth}
        \centering
        \includegraphics[width=\linewidth]{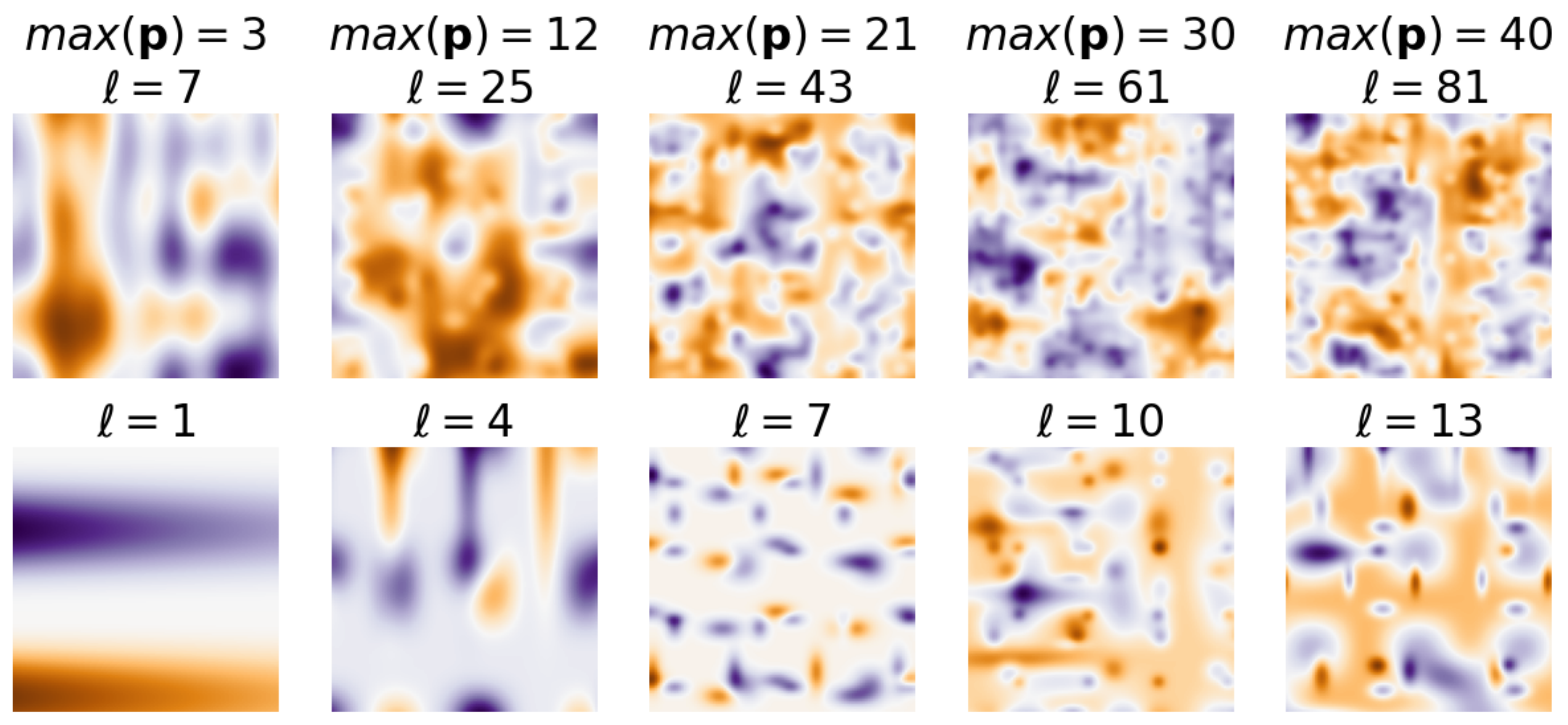}
        \caption{\textbf{Psine: Full and degenerate steering.}  The top rows satisfies $\ell \ge 2\max(\p)+1$ and steers the full space.  The bottom row does not, resulting
        in degenerate filters.}\label{fig:psine}
    \end{minipage}
    \begin{minipage}{.48\linewidth}
        \centering
        \includegraphics[width=\linewidth]{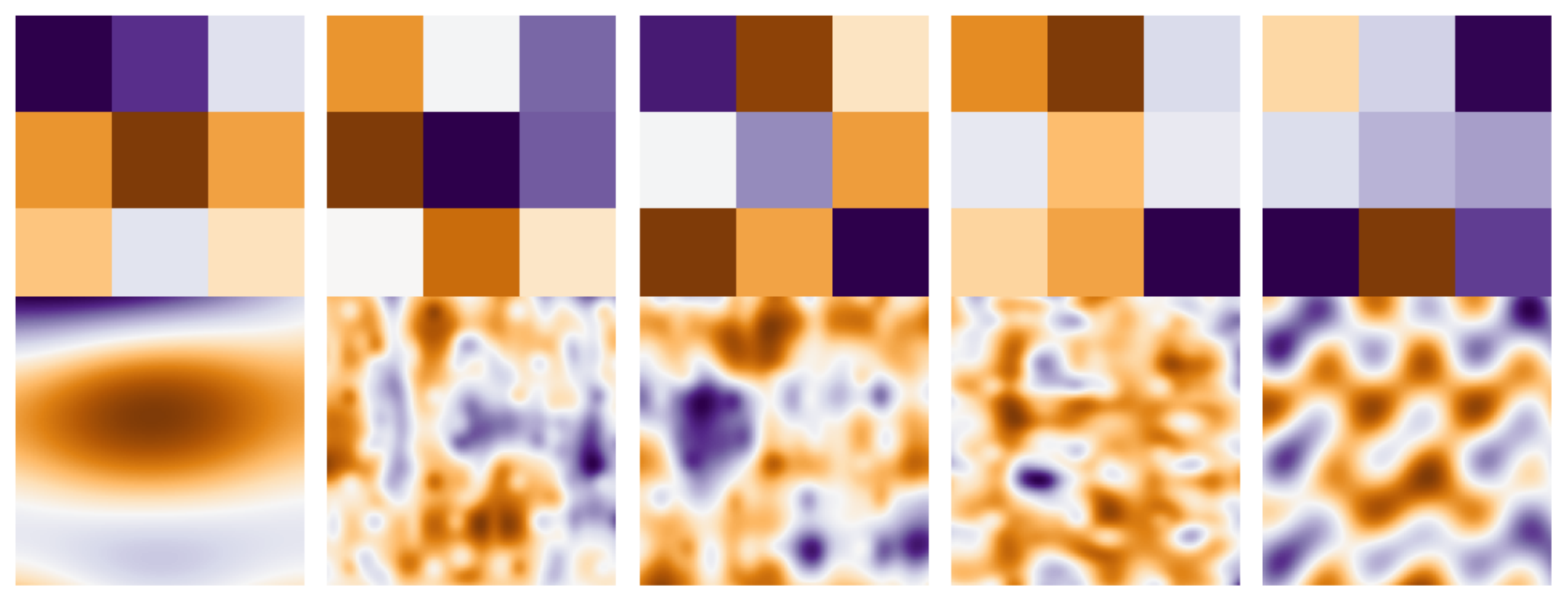}
        \caption{\textbf{Psine filters} visualized as 3$\times$3 filters (top row) and in higher resolution (bottom row). }\label{fig:psine2}
    \end{minipage}\hfill
    \begin{minipage}{.48\linewidth}
        \centering
        \includegraphics[width=\linewidth]{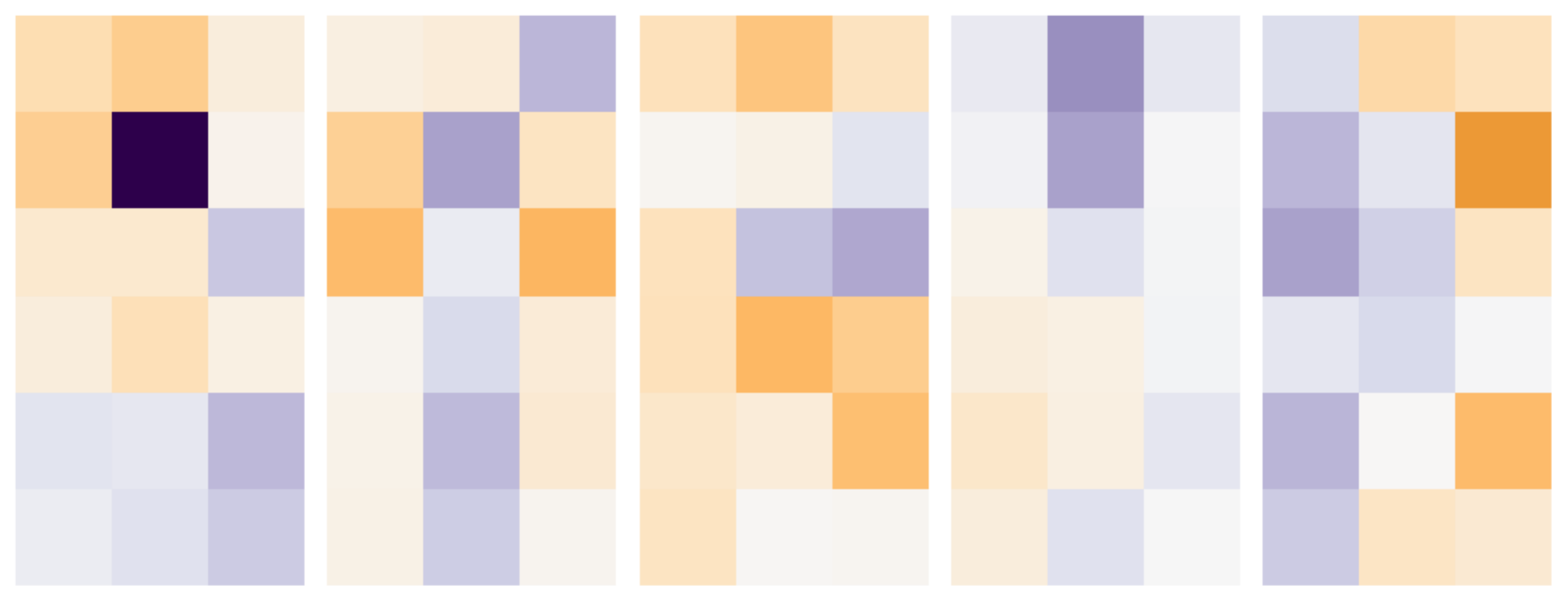}
        \caption{\textbf{GuidedSteer Visualization} using as guide filters a DenseNet121 pre-trained on ImageNet.  The filters do not have a high resolution counterpart.}\label{fig:guidedsteer}
    \end{minipage}
\end{figure}

\textbf{GHaar} is our first steered initialization.  It generalizes and steers Haar wavelet 
basis filters.  A multi-scale Haar transform iteratively downsamples an input image with a box filter to attain approximation images at different scales. It convolves the image at each scale with each of the three basis filters shown in the top row of Fig. \ref{fig:haar}.  We highlight three challenges for a steerable Haar-like initialization: the initialization should have a multi-scale representation; it should work with kernels of shape $3\times3$, $5\times5$ or $h\times w$; the initialization should steer a basis.  Our approach addresses these challenges.

We generalize Haar wavelets by converting the 1-D step functions to sinusoid: $g_{f}(\x) = cos(f \x)$, for a frequency $f$.
  The basis is 1-D separable, as shown in Fig. \ref{fig:haar} and this construction generalizes to any dimensions.
The 1-D vectors $\ba$ and $\bb$ in Fig. \ref{fig:haar} are $\ba = g_{f=0}(\x) = \cos(0 \x)$ and $\bb = g_{f=1}(\x) = \cos(1 \x)$, where $\x = \text{linspace}(0,\pi,m)$ and $m=2$ (top row) or $m=3$ (bottom row).  The basis filters are periodic outside range $f \in [0, 2(m-1))$.  
 Our implementation addresses the three challenges: varying the frequency, $f\in U[0,2(m-1)]$, enables a multi-scale representation;
an outer product with $m=h$ rows and $m=w$ columns constructs $(h,w)$ filters; third, the basis is guaranteed undercomplete.  We steer with only three random frequencies, and the three may not be orthogonal.
Example 2D GHaar filters are visualized in Fig.~\ref{fig:ghaar} for varying frequencies and steering weights, in both high resolution 20$\times$20 filters to give a sense of the filter, and in 3$\times$3 filters. 

\textbf{Psine} steers an undercomplete basis with a polynomial combination of outer products, defined by the polynomial equation $F = \sum_{i=1}^\ell w_i (g(\x_i) g(\y_i)^\top)^{p_i}$.   Psine is a weighted sum of $\ell$ outer products $g(\x_i)g(\y_i)^\top$, where each is raised element-wise to a positive integer power $p_i$, and where $\x_i$ and $\y_i$ are vectors.  
We define $g$ as a sinusoid function.\footnote{Our implementation adopted the GHaar sinusoid with $x=linspace(0,\pi)$ and frequency $f\in U[1,5]$.}  As the number $\ell$ of outer products increases, so does the capacity to represent higher order polynomials.  To relate $\ell$ and $\mathbf{p}$, we note that any $N^\text{th}$ order polynomial is steerable with $2N+1$ basis functions, and it is steerable with as few as $N+1$ basis functions if the function has entirely even or odd order terms (i.e. all terms $x^ny^m$ satisfy $n+m$ is even or odd) \parencite{steerablefilters}.
Thus, when $N$ is too small, the filter cannot be steered in all possible directions.  We will assume $\ell = N$ and secondly that $\p$ contains even and odd positive integers. To steer anywhere in the full space, we define $\ell \ge 2\max(\p)+1$, and we form an undercomplete basis for any given filter by randomly sampling $\ell$ and $\p$.
The filters are whitened to zero mean and unit norm.
Example Psine filters are visualized in Fig.~\ref{fig:psine} for varying $\ell$ and $\max(\p)$, where the top row satisfies and the bottom row does not satisfy the constraint $\ell \ge 2\max(\p)+1$.  Fig.~\ref{fig:psine2} shows example 3$\times$3 filters as well as their higher resolution counterparts.

\textbf{GuidedSteer} initializes spatial filters with the same steering properties as a ``guide'' set.  Without loss of generality of the method, we apply GuidedSteer initialization layerwise to each each spatial convolution layer, where the guide set is the spatial filters of the corresponding layer of an ImageNet Unchanged model with identical architecture.  The primary motivation for layer-wise GuidedSteer is to preserve steering properties of ImageNet trained models in case they contain predictively useful information not captured by GHaar or Psine initializations.  Layer-wise GuidedSteer resets the connections between neighboring pointwise and spatial convolutions while preserving the steering properties within the spatial layer.
More generally, the only restriction on the guide set is the filters have the same kernel shape.  An interesting future application we do not explore but leave to future work is using GuidedSteer to initialize large or wide models from smaller or narrower pre-trained guide models.
  Example GuidedSteer filters are visualized in Fig.~\ref{fig:guidedsteer}.

\textit{Preliminary Definitions and Assumptions:}  Assume a guide collection of $c$ spatial filter kernels of any given shape $(h,w,...)$.   The matrix $G\in \mathbb{R}^{c,m}$ flattens each filter as a row vector.  Without loss of generality, we assume the matrix $G$ represents all same-sized spatial filter kernels in a pre-trained guide deep network.  We wish to obtain a matrix $F \in \mathbb{R}^{n,m}$ of $n$ spatial filters similar to those in $G$.  The number $n$ can be chosen independently of $c$. Assume existence of a certain orthonormal basis, $B \in \mathbb{R}^{m,m}$, with a flattened basis filter per row.  The projection $GB^\top$ represents each filter as a row vector of ``steering'' weights over the basis filters, and each $i^\text{th}$ column describes a distribution $p(w_i|\bb_i)$ over possible steering weights $w_i$ given basis vector $\bb_i$.  A particular choice of basis $B$ allows generating each row of $F$ by independently sampling from the $m$ column distributions $p(w_i|\bb_i)$.  Then, $F$ is a sampled approximation of the guide filters $G$.

We next consider simplifying assumptions to generate the rows of $F$.  First, if we assume that the columns of $GB^\top$ are linearly independent, then each column can be sampled independently.  This is possible via the Singular Value Decomposition (SVD) of $G=USV^\top$, by choosing $B=V^\top$, since the covariance $V^\top G^\top GV$ is diagonalized\footnote{We tested SVD with non-centered $G$ and centered $G-\bar G$ and found no clear winner.} and the columns of $GV$ are linearly independent.\footnote{Linearly independent columns (LIC) follow from the definition of SVD, where $GV = US$ and $U$ and $S$ have LIC.}  In implementation, SVD is memory intensive if $G$ has many rows, so we compute $V$ from the SVD of the covariance $G^\top G = VS^2V^\top$.
Second, we aim to capture the fact that different spatial layers of the network may steer differently.
Therefore, we define a subset of rows $G_\ell \subset G$ in order to define the guide spatial filters of a particular layer $\ell$.  While $B$ is still obtained by SVD of $G$, we define each distribution $p(w_i|\bb_i)$ from columns of $G_\ell B^\top$ rather than $G B^\top$.
Third, we can sample from each $p(w_i|\bb_i)$ independently by assuming a normal distribution $p(w_i|\bb_i) = \mathbb{N}(\mu_i, \sigma_i^2)$ or by Gaussian Kernel Density Estimation (KDE)\footnote{We found no clear winner between KDE and assuming a normal distribution.}.
Finally, to sample a spatial filter: obtain samples $w_i\sim p(w_i|\bb_i)$; concatenate into row vector $\w = [w_1, \dots, w_m]$;
project $\mathbf{w}$ back to the spatial filter domain $\f_i = \mathbf{w}B + \bar G$ where $\bar G = 0$ if using non-centered SVD.  Obtain $F_\ell$ by generating rows $\f_i$.  Without loss of generality, the process repeats for all $\ell$ layers.  Algo. \ref{algo:guidedsteer} is a pseudo-code implementation of layerwise non-centered GuidedSteer.

\subsection{Proposed ExplainSteer Method for Visual Model-based Explanations}\label{methods_explainsteering}
\begin{table}[t]
  \begin{minipage}{.50\textwidth}
  \begin{algorithm}[H]
  \caption{ExplainSteer Algorithm.}\label{algo:explainsteer}
  \small
  \begin{algorithmic}[1]
  \Procedure{ExplainSteer}{$F, B, \w$}
    \State Inputs: $F\in\mathbb{R}^{n,m}, B\in\mathbb{R}^{m,m}, \w\in\mathbb{R}_{\ge0}^{n}$
    \State $\e^{(d)} = \w \left|F B^\top\right|$
    \State $\e^{(1)} = [\e^{(1)}_h, \e^{(1)}_w, \dots] = \text{zeros}(h), \text{zeros}(w), \dots$
    \For{$e_j \in \e^{(d)}$}
      \State $h_\text{idx}, w_\text{idx}, \dots = \text{get1dBasisVectorIdx}(j)$
      \State $\e^{(1)}_h[h_\text{idx}] \mathrel{+}= e_j^{\frac{1}{d}} \;;\; \e^{(1)}_w[w_\text{idx}] \mathrel{+}= e_j^{\frac{1}{d}} \;;\; \dots$
    \EndFor
    \State $E = \text{asMatrix}(\e^{(1)}, \text{shape=}(d, \max(h,w,\dots)))$
    \State $\e^{(0)} = E\text{.mean(0)}$
    \State \textbf{return} $\e^{(d)}, \e^{(1)}, \e^{(0)}$
    \EndProcedure
  \end{algorithmic}
  \end{algorithm}
  \end{minipage}%
  \hfill
  \begin{minipage}{.49\textwidth}
\begin{table}[H]
  \centering
\caption{ExplainSteer Notation.}
\small
  \begin{tabular}{ r | l}
    \toprule
    Variable & Description \\
    \midrule
    $d \ge 1$ & Dimensionality of the basis\\& and spatial filter kernels. \\
    $m \ge 1$ & Size of the spatial filter kernel\\& and equivalently the basis size.\\
    $F \in \mathbb{R}^{n,m}$ & Matrix of spatial filter kernels.\\
    $B^{(d)} \in \mathbb{R}^{m,m}$ & Matrix describing the basis,\\& one basis vector per row.\\
    $\w \in \mathbb{R}^n$ & Saliency weights, one per\\& spatial filter kernel.\\
    $\f \in \mathbb{R}^m$ & Flattened spatial filter kernel\\& (a row of $F$).\\
    $\bb_i^{(d)} \in \mathbb{R}^{m}$ & Flattened basis filter\\& ($i^\text{th}$ row vector of $B^{(d)}$).\\
    $\otimes$ & Tensor outer product operator.\\
    $\circ$ & Element-wise multiplication.\\
    \bottomrule
  \end{tabular}
  \label{table:explainsteer_variables}
\end{table}
  \end{minipage}
\end{table}
To better understand how spatial filters are steered, sanity check our initializations, and expose inefficiencies in deep networks, we develop visual model-based explanations.  Our explanations differ from the typical usage of explainability in that we do not explain model outputs but rather the internal properties of the model itself.  The ExplainSteer method offers a spectrum characterizing relevant frequencies and edge orientations of a given set of spatial filters, two human-interpretable dimension reductions of the spectrum, and a saliency method.  The explanation utilizes linear projection with any interpretable basis.  We primarily adopt the DCT-II basis because it is 1-D separable and relates to the DCT2, GHaar and Psine initializations.  
We interpret ExplainSteer results in Sec. \ref{results_explainability} to show how the visual explanations verify an initialization and expose deep network inefficiencies.

Preliminary Definitions: Table \ref{table:explainsteer_variables} summarizes most important notation.  Assume no notation from previous sections is used in this section.  A $d$-dimensional spatial filter is a tensor.  Given a set of $n$ spatial filters of dimension $d$, we define each row of a matrix $F \in\mathbb{R}^{n,m}$ as a spatial filter $\f'\in \mathbb{R}^{h,w,\dots}$ flattened to a row vector $\f \in \mathbb{R}^m$. We similarly define an orthonormal DCT-II basis $B^{(d)} \in\mathbb{R}^{m,m}$ with flattened basis filters $\bb_i^{(d)}$ on rows.  For any $d\ge1$, the basis decomposes as a tensor outer product of the 1-d basis vectors, $\bb_i^{(d)} = \text{flatten}(\bb_1^{(1)} \otimes \dots \otimes \bb_j^{(1)} \otimes \dots \otimes \bb_d^{(1)})$ where each $\bb_j^{(1)} \in \mathcal{B}_{N_j}^{(1)}$, and
\begin{align}
  \mathcal{B}_{N_j}^{(1)} = \Bigg\{ &s\cos\left(\frac{2\pi}{N_j}(k+0)(\x+.5)\right) \sVert[4] 
                            s=\begin{cases}
                              \sqrt{\frac{1}{N_j}} &\text{\small{if }}k=0\\
                              \sqrt{\frac{2}{N_j}} &\text{\small{if }}k\ne 0
                            \end{cases},
                            & k\in [0..N_j-1],
                            \x = [0..N_j-1] \Bigg\}
\end{align}
where $j$ indexes the $d$ spatial dimensions, $N_j$ is the size of the spatial filter and size of the basis in dimension $j$, $k$ determines the frequency of a basis vector, and $s$ ensures orthonormality. 
In this study we consider 2-$d$ basis filters.  It is a matrix with spatial dimensions of $h$ rows and $w$ columns, and it has $m=hw$ elements.  It is an outer product of 1-$d$ basis vectors $\bb_1^{(1)} \in \mathcal{B}_{h}^{(1)} \subset \mathbb{R}^h$ and $\bb_2^{(1)} \in \mathcal{B}_{w}^{(1)} \subset \mathbb{R}^w$. See Fig. \ref{fig:haar} for 2-$d$ visual examples when $m=2^2$ (top row) or $m=3^2$ (bottom row); they are correct up to scale $s$.  Appendix \ref{appendix_DCT-II} visualizes DCT-II bases for $d=2$.

Columns of the linear projection $P = FB^\top$ have nice interpretation.  Each column corresponds to a basis filter $\bb_i$.  The total column magnitude characterizes how much energy each basis filter $\bb_i$ contributes to $F$.  We define an energy spectrum $\e^{(d)} = \begin{bmatrix}e_1^{(d)} & \dots & e_m^{(d)} \end{bmatrix}$, where each $i^\text{th}$ column of $P$ is reduced via a weighted sum over its rows $j$ (note that $j$ in $\f_j$ below and $\bb_j$ above have different meaning):
\begin{align} e_i^{(d)} = \sum_j w_j \left| \f_j \left(\bb_i^{(d)}\right)^\top \right| \label{eq:explainfreq_energy}\quad\implies\quad
  \e^{(d)} = \w \left| FB^\top \right| 
\end{align} 
Larger $e_i^{(d)}$ corresponds to larger importance of basis filter $\bb_i^{(d)}$ to the filters $F$.  For example, when $F$ represents 2$\times$2 spatial filters (i.e. $m=2^2$, $d=2$), it has an energy spectrum $\e^{(d)}$ of four values corresponding to the importance of vertical, horizontal, diagonal and bias components in Fig. \ref{fig:haar}.  The weight $w_j$ can introduce selective bias towards important spatial filters, or it can be $w_j=\frac{1}{n}$.

To visualize and verify an initialization method, we use the default weight, and we compute and visualize the energy spectrum for each layer of a model.
In order to expose inefficiencies in deep networks, we also visualize the spectra by setting $w_j$ with weight saliency.  
We visualize the energy spectra of various deep networks, both with and without saliency in Appendix \ref{appendix:saliency_vs_not}.  Detail describing the visualization is in Sec. \ref{results_explainability}.  We observe that saliency makes the explanations more similar for a given architecture, thus making the explanation less dependent on initialization and more descriptive of the architecture.  We define each saliency weight similarly to an input times gradient method method, except we use weights rather than pre-activations, and we sum over several examples:
\begin{align}
  w_j = \sum_{m}^M \mathbf{1}^\top \left|  \sum_{(\x_i,\y_i) \in \mathcal{M}_m} \od{\left(\y_i^\top \frac{\hat \y_i}{\hat\y_i'}\right)}{\f_j} \circ \f_j \right|,
  \label{eq:saliency}
\end{align}
where $\mathcal{M}_m$ is a minibatch containing labeled datapoints $(\x_i, \y_i)$ with the ground truth labels $\y_i$ in $[0,1]$; $M$ is the total number of minibatches\footnote{We set $M=15$ or $M=50$ with a minibatch size of 4.}; $\hat \y_i$ is the model output and $\hat\y_i'\triangleq \hat\y_i$ is a constant that rescales gradients so each output has even contribution; the absolute value is element-wise over the $m$ weights in the spatial filter $\f_j$; $\circ$ is element-wise multiplication; and $\mathbf{1} \in \mathbb{R}^m$ sums across all $m$ spatial positions in $\f_j$.

Back-projecting $\e^{(d)} \in \mathbb{R}^m$ onto the 1-D basis to obtain $\e^{(1)} \in \mathbb{R}^{h+w+\dots}$ or $\e^{(0)} \in \mathbb{R}^{\max(h,w,\dots)}$ is an interpretable dimension reduction.  Each scalar $e_\ell^{(1)}$ denotes an energy of a frequency in a spatial dimension (e.g. ``vertical components have low frequency''), and $e_q^{(0)}$ describes the energy of the $q^\text{th}$ frequency generally (e.g. ``the spatial filters have low frequency components'').
Therefore, the dimension reduction makes the spectrum easier to explain.  Consider that $5\times5$ spatial filters have $m=25$ elements, $m=25$ basis filters, and correspondingly $\e^{(d)}$, $\e^{(1)}$ and $\e^{(0)}$ give $25$, $10$ and $5$ values.

We next describe how to perform the reduction.
Consider a 2-d spatial filter of $m$ elements, so we have $m$ basis filters, $d=2$, and $\f = w_1 \bb_1^{(d)} + \dots + w_m \bb_m^{(d)}$.  A clarifying note on notation: $w_{1\dots m}$ are steering weights; the saliency weight variables $w_j$, defined above, are not explicitly used here.  Each 2-d basis filter in the sum can be decomposed as a tensor outer product of 1-d basis vectors $w_i \bb_i^{(2)} = w_i \bb_i^{(d)} = w_i(\bb_a^{(1)} \otimes \bb_b^{(1)}) = \sqrt[d]{w_i} \bb_a^{(1)} \otimes \sqrt[d]{w_i}\bb_b^{(1)}$, where the steering weight $w_i$ is equally distributed.  
We next substitute Eq. \ref{eq:explainfreq_energy}, $w_i \triangleq e_i^{(d)}$. Finally, for each distinct 1-d basis vector in the tensor outer product, we obtain the corresponding back-projected energy $e_\ell^{(1)}$ in Eq. \ref{eq:explainfreq_backproject}.  The expression attributes part of an energy $e_i^{(d)}$ to a particular 1-d basis vector $\bb_\ell^{(1)}$ only if it was used to construct the corresponding basis filter, $\bb_i^{(d)}$.
\begin{align}
  e_\ell^{(1)} &= \sum_i^m \delta_{i,\ell} \sqrt[d]{w_i} = \sum_i^m \delta_{i,\ell} \sqrt[d]{e_i^{(d)}} \label{eq:explainfreq_backproject}
               \qquad\text{where}\qquad\delta_{i,\ell} = \begin{cases}
  1 & \text{if }\bb_i^{(d)} \text{ is constructed with } \bb_\ell^{(1)}\\
  0 & \text{otherwise}
\end{cases}
\end{align}

Eq. \ref{eq:explainfreq_backproject0} obtains $\e^{(0)}$ by reshaping $\e^{(1)}$ into a matrix $E^{(1)} \in \mathbb{R}^{d,\max(h,w,\dots)}$.  
Each column of $E^{(1)}$ corresponds to a frequency of the DCT-II 1-d basis $\mathcal{B}_{N_j}^{(1)}$.
The reshaping to $E{(1)}$ and reduction of $\e^{(1)}$ to $\e^{(0)}$ is straightforward when all $j$ spatial dimensions have equal shape (i.e. $h = w = \dots$).  In this case, 
$\mathcal{B}_{N_j}^{(1)} = \mathcal{B}_h^{(1)} = \mathcal{B}_w^{(1)} = \dots$ since $N_j$ would be the same for all $j$, and a column sum or column mean of $E^{(1)}$ gives per-frequency energies $\e^{(0)} \in \mathbb{R}^d$. 
We do not analyze $\e^{(0)}$ when spatial dimensions have different sizes $N_j$, since (a) it would be unclear how to construct $E^{(1)}$, and (b) deep networks typically use square kernels (such as $3\times 3$ or $5\times 5$).
\begin{align}
  \e^{(0)} &= \frac{\mathds{1} E^{(1)}}{d} \qquad\text{where}\qquad \label{eq:explainfreq_backproject0}
  E^{(1)} = \begin{bmatrix}
    \text{($h$ energies in $\e^{(1)}$ for dimension 1)}\\
    \text{($w$ energies in $\e^{(1)}$ for dimension 2)}\\
    \dots
  \end{bmatrix}
\end{align}

Algo. \ref{algo:explainsteer} summarizes this explanation method.
In Sec. \ref{results_explainability}, we analyze ExplainSteer explanations of different network architectures.  Our analysis leads to saliency-based pruning of fixed filter networks in Sec. \ref{sec:results_pruning}.

\subsection{Proposed ChannelPrune Method for Smaller and Faster Models}\label{sec:pruning}

Motivated by our analysis of ExplainSteer visual explanations in Sec. \ref{results_explainability}, we developed ChannelPrune to remove unnecessary spatial convolution filters from deep networks.  We treat a deep network as a directed graph and then also carefully remove parameters throughout the network that are connected to unnecessary spatial convolution filters.  The result is smaller, more efficient networks. The method has five steps: (1) obtain a saliency score for each spatial filter kernel; (2) zero out the least salient spatial filter kernels globally across the network; (3) prune entirely zeroed input and output channels of the spatial convolutions layers; (4) remove channels from neighboring layers by a graph traversal to correct inconsistencies in the network; (5) re-initialize any remaining zeroed kernel values with our FillZero method (described below).  ChannelPrune preserves the most salient steering properties of the input model.

The saliency score in Eq. \ref{eq:saliency} can serve for step (1). We found that even using just the absolute value of the gradients is sufficient (i.e. assume $\f$ is a ones vector after computing the gradient).  Step (2) uses saliency to zero out least salient spatial kernels instead of randomly choosing them.  The results of our experiment to sanity check the saliency method in Appendix \ref{appendix:sanity_check_saliency} and Fig. \ref{fig:sanity_check_saliency2} validate this choice, since the results on spatially fixed networks imply that randomly removing important fixed filters will strongly and negatively impact performance.  Step (3) removes input and output channels from sparse spatial convolution layers by using the definition from Eq. \ref{eq:steerability} that each output channel of a spatial convolution is a sum of multiple cross-correlated input channels.   Given a particular output channel $\bO_o$, if all corresponding kernels $\f_{o,i}$ were zeroed in step two, then that output channel $\bO_o$ and all connected kernels $\f_{o,i}$ can be removed.  Similarly, given an input $\I_i$, if all corresponding kernels $\f_{o,i}$ were zeroed, then that input channel $\I_i$ and its connected $\f_{o,i}$ can be removed.  Appendix \ref{appendix:table_pruning_zeroing} details why saliency-based channel pruning cannot typically remove all zeroed filters.  After pruning a spatial layer, the network is inconsistent and needs correction. Step (4) ensures neighboring convolution, BatchNorm and Linear layers are channel pruned to precisely match the pruned spatial convolution layers. Step (4) represents the network as a graph and utilizes a graph traversal.  We keep track of which specific channels were pruned, the direction traveled, the previous node, and we implement specific handling of various kinds of connections between layers.  We refer to our source code for a complete reference \parencite{our_source_code_github}.  Finally, in step (5), we re-initialize all spatial weights that are close to zero with FillZero: entirely zero spatial kernels are re-initialized with Kaiming uniform random values; remaining zero weights are re-initialized with Gaussian noise of mean and variance equal to the spatial filter's mean and variance.  We found it necessary to double the learning rate of pruned networks in order to fine-tune the non-fixed (non-spatial) weights.  Sec. \ref{sec:results_pruning} and Appendix \ref{appendix:comp_savings_same_acc} show that fixed pruned models generally have matching predictive performance to the fully learned baseline while being both smaller and faster.

\section{Experiments and Explanations}\label{results}

\textbf{Experiment Design:} We conduct ablative experiments in Sections \ref{results_chexpert}, \ref{sec:compsave}, and \ref{results_robustness} verifying our \fixedFilters design principle that all spatial convolution kernels can be fixed at initialization, that the fixed spatial initialization should have a steered representation, and that fixed models give speed and accuracy gains.  In Sec. \ref{results_explainability}, we analyze ExplainSteer visual explanations, and we exploit these explanations in Sec. \ref{sec:results_pruning} to verify, in Sec. \ref{sec:results_pruning_subsection}, our \nimbleness design principle.  We show that most spatial weights, and indeed most weights generally, are not necessary for inference nor training.  \ExplainFix models have fixed spatial weights.  They are always faster, smaller with ChannelPrune pruning, and as accurate as their corresponding fully learned baselines.

\textbf{Datasets:} CheXpert \parencite{dataset_chexpert} is a chest x-ray dataset of 223k:235 images for classification of five pathologies.  BBBC038v1 \parencite{dataset_BBBC038v1} contains 670:65 annotated microscopy images for semantic segmentation.  Appendix \ref{appendix_dataset_details} details the datasets and how we split, pre-process and evaluate them.

\textbf{Models:} We implement 13
baseline models using four
distinct kinds of deep network architectures.  The DenseNet architecture \parencite{densenet} uses $1\times 1$ and $3\times 3$ convolutions, ResNet \parencite{resnet} uses the bottleneck ($1\times1$, $3\times 3$, $1\times 1$), EfficientNet \parencite{efficientnet} uses depthwise separable bottlenecks where most weights are pointwise convolutions, and U-NetD is our U-Net \parencite{unet_ronneberger2015} implementation with entirely depthwise separable bottlenecks and channel expansion from MobileNetV2 \parencite{mobilenetv2}.  We evaluate each architecture with non-spatial weights initialized to random values (``FromScratch'') or ImageNet values (``PreTrained'').  Table \ref{table:model_num_params_and_savings} and Table \ref{table:bbbc_num_params} highlight how many spatial filter parameters each architecture has.  See Appendix \ref{appendix_model_details} for extensive details.

\textbf{Reproducibility:} Hyper-parameter configuration is defined in Appendix \ref{appendix_hyperparameter_details}. Source code at \parencite{our_source_code_github}.

\subsection{Steered Fixed Spatial Initializations Improve Predictive Performance}\label{results_chexpert}

\begin{figure}[t]
  \begin{minipage}{.49\textwidth}
    \includegraphics[width=\linewidth]{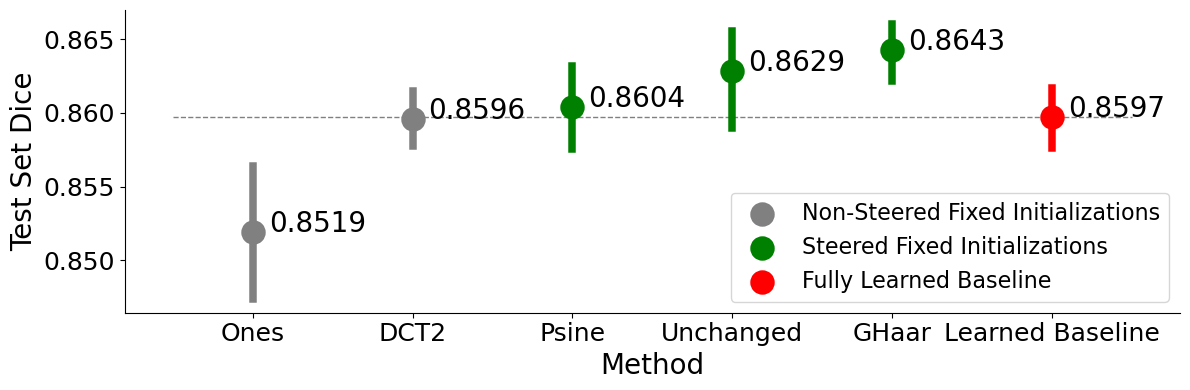}
    \caption{
      \textbf{Steering Helps.}  Steered fixed models significantly outperform the non-steered models.  GHaar significantly outperforms the learned baseline.  Evaluated on BBBC038v1.
    }\label{fig:BBBC_winning_ticket2}
  \end{minipage}\hfill
  \begin{minipage}{.49\textwidth}
    \includegraphics[width=\linewidth]{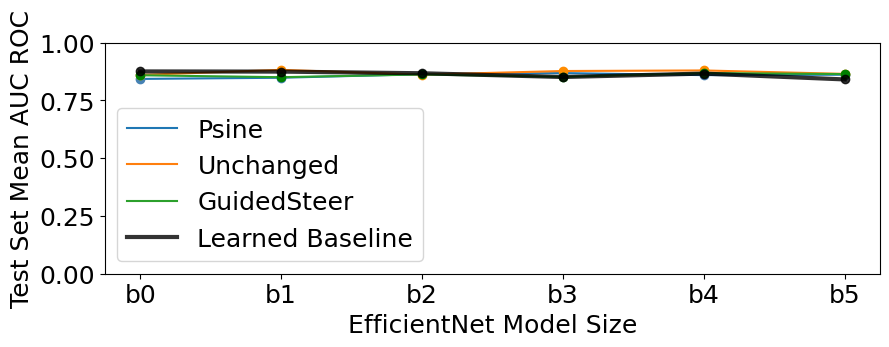}
    \caption{
      \textbf{\ExplainFix is Robust} to training larger models.
    }\label{fig:CM8_robust_overfitting}
  \end{minipage}
  \end{figure}
  \textbf{BBBC038v1:} Steered initializations improve performance of spatially fixed models.  Fig. \ref{fig:BBBC_winning_ticket2} shows that the steered initializations significantly outperform the non-steered initializations.  It also shows the GHaar initialization significantly outperforms the learned baseline.  The figure represents 36 independently trained models, with six models per method.  For each method on the x axis, we select the best of the six corresponding models and show its average test set performance for epochs 151-300 with a 95\% confidence interval.  Each model was trained for 300 epochs and evaluated on the out-of-distribution dataset at each epoch.  We show average performance per epoch of all 36 models in Appendix \ref{appendix:BE11_in_vs_out_of_distribution}, and we compare it to a replicate the study with another set of 36 models evaluated the in-distribution dataset.  Both appendix figures confirm the results in Fig. \ref{fig:BBBC_winning_ticket2} to show that, in general, all steered spatially fixed models approximately equal or outperform the learned baseline with statistical significance.

\begin{figure}[t]
  \centering
  \begin{minipage}{.7\textwidth}
  \subfloat[CheXpert Ablative Results.\label{fig:chexpert_C8}]{
    \includegraphics[width=\textwidth]{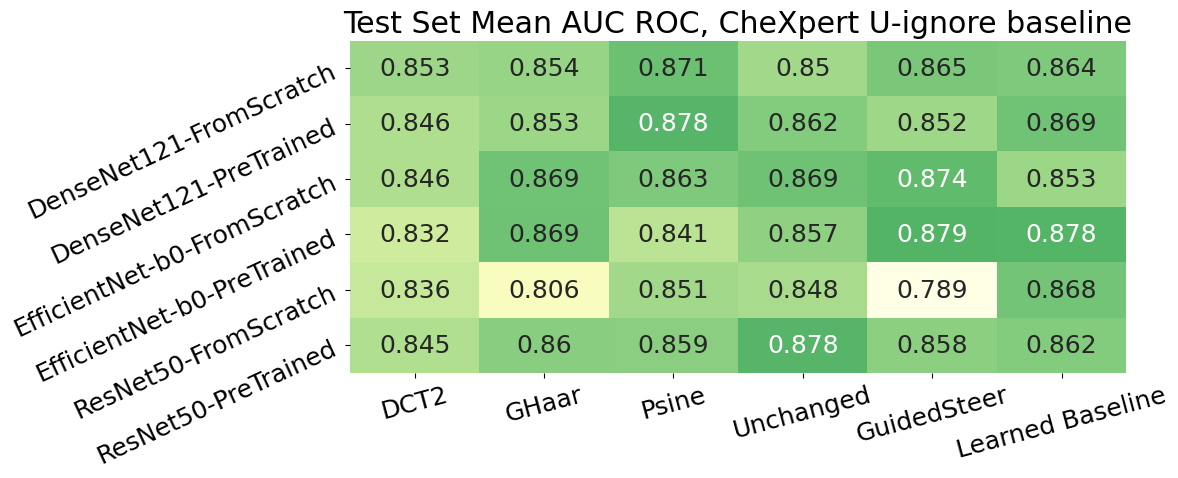}%
  }
  \end{minipage}
  \begin{minipage}{.29\textwidth}
  \subfloat[Significance tests.\label{table:chexpert_C8}]{
    \footnotesize
    \begin{tabular}{ll}
\toprule
     Method &  Equal? (p) \\
\midrule
       DCT2 &  \cellcolor{red!25}NO (2e-05) \\
     Ghaar & \cellcolor{green!25}YES (0.026) \\
      Psine &  \cellcolor{green!25}YES (0.31) \\
  Unchanged &  \cellcolor{green!25}YES (0.33) \\
GuidedSteer &   \cellcolor{green!25}YES (0.3) \\
\bottomrule
\end{tabular}

  }
  \end{minipage}
  \caption{\textbf{Matching Accuracy.}  Steered fixed filter models approximately equal the baseline.  Only DCT2 is not steered, and it has lower performance. Evaluated on CheXpert.}
\end{figure}

\textbf{CheXpert:} Results show that (a) steered fixed filter methods significantly outperform the non-steered DCT2, (b) steered methods 
outperform the fully learned baselines on specific architectures, and (c) across all types of models generally, the steered fixed filter models approximately equal the baseline.  Fig. \ref{fig:chexpert_C8} represents 36 independently trained models, each number represents the test set average ROC AUC
over five tasks, and each column represents 6 trained models.
The table in Fig. \ref{table:chexpert_C8} reports the two-tailed p-values of paired-sample Wilcoxon tests to evaluate each column of the corresponding figure.  The test supports our findings (a) and (c).   On specific architectures, the figure suggests that Psine initialization is best for DenseNet, GHaar and GuidedSteer are best for EfficientNet, and in Appendix \ref{appendix_effects_of_pretraining} we show that ImageNet Unchanged outperforms the pre-trained ResNet50 baseline on our hold out set of 56k validation images consistently for all epochs.

\subsection{Computational Savings:}\label{sec:compsave}

\begin{table}[t]
  \centering
  \caption{\textbf{\ExplainFix Saves Time.}}\label{table:model_num_params_and_savings}
  \small
  \begin{tabular}{llll}
\toprule
\textbf{Architecture} & \textbf{Spatial Params} & \multicolumn{2}{c}{\textbf{SPE\textsuperscript{*}}} \\
               &Num (\% total) & ExplainFix & Baseline\\
\midrule
ResNet50        & 11.3e6 (48\%) & 191 (\textbf{17\%}) & 231 \\
DenseNet121     & 2.1e6 (31\%) & 347 (\textbf{9\%})  & 383 \\
EfficientNet-b0 & 0.2e6 (5\%) & 259 (\textbf{3\%})  & 267 \\
\bottomrule
\end{tabular}

  \\\footnotesize  (*) SPE compares Mean \underline{S}econds \underline{P}er Training \underline{E}poch of the \ExplainFix and Baseline models and reports \textit{(\% savings)} with respect to the baseline.
  SPE of ResNet and DenseNet also analyzed in Fig. \ref{fig:compute_efficiency2}.
\end{table}
\begin{figure}[t]
  \centering
  \includegraphics[width=.6\linewidth]{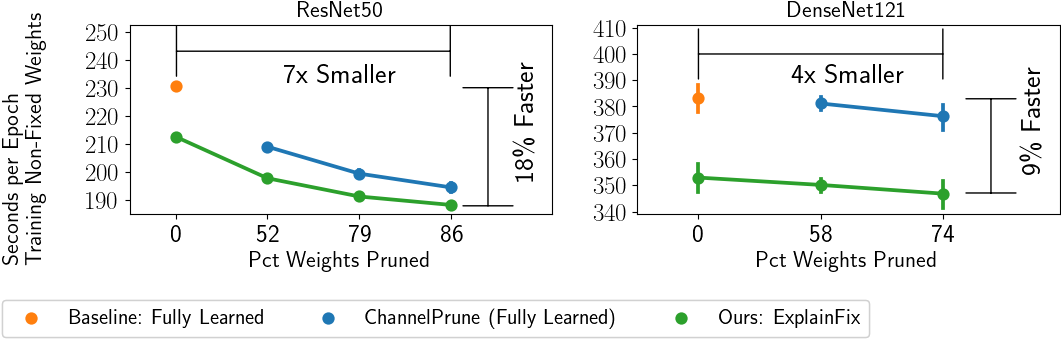}
  \caption{\textbf{Always Faster And Smaller} than fully learned baselines.}\label{fig:compute_efficiency2}
\end{figure}
By design, fixed spatial filter models use less computation during training because the spatial filter weights are not updated during backpropagation.  
Pruning fixed filter models results in further savings.  Table \ref{table:model_num_params_and_savings} shows how our \ExplainFix models reduce per-epoch training time.
The ResNet50 architecture has the largest percent of spatial parameters and shows the best savings from \ExplainFix.  The \ExplainFix ResNet50 model is 5x smaller (only 21\% of its parameters were kept after channel pruning) and 17\% faster when training non-spatial weights.  The DenseNet121 model is similarly pruned and fixed, while the EfficientNet model is fixed.  We show expanded results in Fig. \ref{fig:compute_efficiency2} that \ExplainFix models are always faster than the fully learned baseline, even when the baseline is also pruned.
Finally, to ensure integrity of our results, we sanity check the predictive performance of our methods:
Sec. \ref{results_chexpert} verified that spatially fixed models approximately equal or surpass baseline performance;
Appendix \ref{appendix:comp_savings_same_acc} shows that fixed and pruned \ExplainFix models match baseline predictive performance while always being faster and smaller.

All SPE measurements in the Table \ref{table:model_num_params_and_savings} and Fig. \ref{fig:compute_efficiency2} were 
obtained by training each model on the same isolated server and NVIDIA RTX2080 Ti GPU for 20 epochs, with no other sources of computational or input-output bottlenecks.  \ExplainFix models are always faster than their fully learned counterparts.

\subsection{Steered and Fixed Initializations Improve Robustness}\label{results_robustness}

\textbf{Steering is important to spatial filter initialization}.  We observed on both datasets that steered initializations significantly improve performance over non-steered initializations.  The results suggest that relying on architecture and learning dynamics alone to steer fixed spatial feature maps causes lower performance.  When the fixed spatial filter kernels are initialized with a redundant and steered representation, performance typically increases.

\textbf{Better robustness to larger learning rate:} On BBBC038v1, we evaluate whether the fixed filter models are sensitive to an increase in learning rate.  In Appendix \ref{appendix:BE10_results}, we train and evaluate 36 models (six models per initialization method) with learning rate 0.008, and we replicate the experiment with learning rate 0.02.  We observe the larger learning rate models more clearly outperform the baseline early in training, and that the non-steered DCT2 method performs significantly worse.  Non-steered spatial initializations rely on the learning dynamics to build steerability into the network architecture.  When these dynamics are dysfunctional, such as induced by larger learning rate, the model struggles to learn.  These results suggest steered fixed initialization improves robustness to an increased learning rate, and further that steered initialization facilitates learning.

\textbf{Robust to varying model capacity:} On CheXpert, we evaluate how performance changes with varying model size, all other parameters constant.  In Fig. \ref{fig:CM8_robust_overfitting} we observe that fixed filter models do not underperform when the model capacity increases.  The x-axis corresponds to models of increasing capacity via compound scaling; \texttt{b0} has smallest depth and channel width, and fewest number of learned parameters while \texttt{b5} has 7x more parameters.  Each point represents the test set performance of a model trained for 80 epochs on CheXpert.  We also evaluate smaller models by pruning the baseline to as little as 10\% of its original size.  In Fig. \ref{fig:compute_eff2_acc}, we observe that pruned fixed models have approximately equal performance to pruned learned models.  In Sec. \ref{sec:pruning}, we describe the pruning method.
The results suggest that fixed spatial filter models behave approximately equally to the learned model as capacity increases or decreases.

\subsection{ExplainSteer: Explanations of Spatial Filter Steering}\label{results_explainability}

Our ExplainSteer explanations are primarily visual, applied globally to all spatial filters for $\e^{(0)}$ and layerwise as a heatmap for $\e^{(2)}$.  The explanations highlight inefficiencies of deep networks and led us to the \nimbleness principle that nearly all spatial filters are unnecessary for both inference and training. 

\begin{figure*}[t]
  \centering
  \subfloat[Per-Layer Spectra $\e^{(2)}$\label{fig:explainsteer_spectra}]{
    \FigExplainSteer
  }
  \subfloat[Per-model Spectra $\e^{(0)}$: Low Frequencies (Green) Dominate\label{fig:explainsteer_e0}]{
    \includegraphics[height=2.5cm]{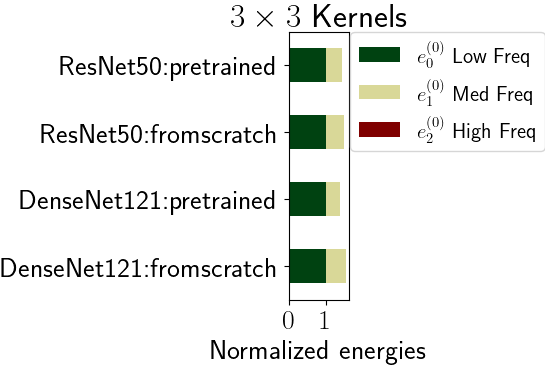}
    \includegraphics[height=2.5cm]{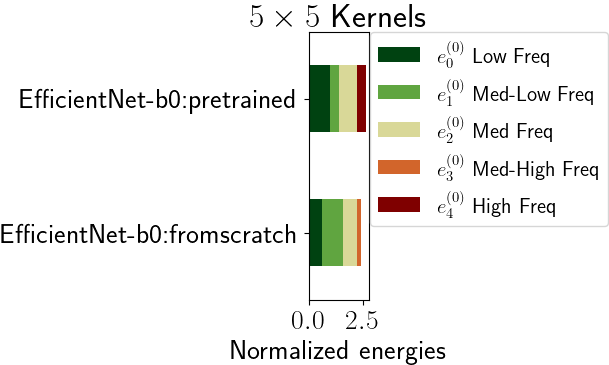}
    }
    \captionof{figure}[c]{\textbf{Low Frequencies and Early Layers Suffice.}  \textbf{(a)} The $\e^{(2)}$ spectra for DenseNet121 exposes its inefficiencies as vertical, horizontal and repeating trends.  \textbf{(b)} The $\e^{(0)}$ spectra shows that spatial filters across all models tested prioritize low frequencies (green) over high frequencies.  More examples in Appendix \ref{appendix:saliency_vs_not}.}
\end{figure*}

\textbf{\textit{Observation 1:} Spatial kernels prioritize low frequencies.}
Fig. \ref{fig:explainsteer_e0} shows $\e^{(0)}$ spectra globally across different models.  The models show a preference for lower frequencies, especially in ResNet50 and DenseNet121.  
Concurrent with our work, \parencite{tancik2020fourier} showed that multi-layer perceptrons have a natural bias towards low frequencies.  Our saliency-based visual explanations suggest spatial convolution layers in CNNs may share this bias.  Since saliency is dependent on a dataset, we also note a relation to the dataset domain.  In chest x-ray data, low frequencies correspond to gradients and lines of structural features, like the outlines of the clavicle, diaphram, heart and ribs.  These strong features anchor the key focus areas and are necessary anatomical landmarks to address the primary tasks in the CheXpert dataset. Thus, it also makes sense on CheXpert that low frequencies would be most salient.

\textbf{\textit{Observation 2:} Nimbleness Insight.}  Deep Networks tend to prioritize low frequency spatial convolution filter kernels located at early layers of the model.  High frequency kernels and other spatial layers are largely unnecessary.
Fig. \ref{fig:explainsteer_spectra} visualizes an example $\e^{(2)}$ explanation.  Appendix \ref{appendix:saliency_vs_not} interprets many of these plots with varying model architecture and initialization.
Each row of the heatmap visualizes energies of a single basis filter over all spatial convolution layers of the network.
Each column shows the $\e^{(2)}$ energy spectrum of the spatial filter kernels in a given spatial convolution layer.
The bar plots show the respective row and column sums of the heatmap.
The basis vector index on y-axis corresponds to basis filter index numbers shown in Appendix \ref{appendix_DCT-II}; indices sort the filters from lowest to highest frequency.  In the plots, a column of 9 and 25 values correspond, respectively, to the $3\times 3$ and $5\times 5$ basis vectors shown and numbered in Appendix Fig. \ref{fig:DCTII_bases3} and \ref{fig:DCTII_bases5}.

The energy spectra across layers of the heatmap have intuitive interpretation.  For instance, the relative contribution of bias, vertical gradients, and horizontal gradients are visualized as the first three heatmap rows (basis filter indices zero, one, and two). For any given heatmap column, a vertical trend from dark to bright shows that the filters prioritize low frequency components more than high frequency components.  For any given row, a horizontal trend from dark to bright shows that the earlier filters have higher energy than later filters.  Repeating horizontal trends in Fig. \ref{fig:explainsteer_spectra} of a DenseNet121 model suggest the spatial convolution layers towards the end of dense blocks are unnecessary.
The additional ExplainSteer heatmaps in Appendix \ref{appendix:saliency_vs_not} 
show these trends consistently across architectures and initializations.   These explanations also enable comparison of pre-trained versus random initialization in fully learned baselines.  We observe that ImageNet pre-training concentrates, or sparsifies, energy onto low frequency components and fewer spatial layers better than learning from random networks.  
The main observations of the nimbleness insight are that the deep networks analyzed are inefficient by design.

\textbf{\textit{Observation 3:} ExplainSteer enables sanity checks on the initialization method.}  The heatmaps without saliency describe how a model is steered.  The DCT2 initialization, for example, should appear as a flat (green) heatmap.  The GuidedSteer initialization should look exactly like its guide model, ImageNet Unchanged.  In Appendix \ref{appendix:saliency_vs_not}, we present these sanity checks and we observe that all sanity checks pass our visual inspection.  The visual is also useful to verify a fixed initialization was indeed fixed during training.

\subsection{Exploiting ExplainSteer Explanations}\label{sec:results_pruning}
The ExplainSteer explanations suggest that all models we analyzed are inefficient, and that different fixed initializations can concentrate the ExplainSteer energy spectra in sparse locations.  To test and exploit this explanation, we propose to eliminate nearly all kernels in spatial convolution layers with three sets of experiments.  The first two experiments zero out spatial kernels.  The third experiment prunes networks by removing zeroed input and output channels systemically across the network.
Sec. \ref{sec:results_unnecessary_inference} shows that nearly all spatial kernels are unnecessary for inference.  Sec. \ref{sec:results_unnecessary_training} shows that most spatial kernels are unnecessary for training non-spatial weights.  Sec. \ref{sec:results_pruning_subsection} shows that deep networks can be fixed and pruned to a fifth their size with approximately equal predictive performance to a fully learned baseline.

\subsubsection{Few Spatial Filters Suffice for Inference}\label{sec:results_unnecessary_inference}
Least salient kernels are unnecessary for inference and for training.  We conduct two sub-experiments to verify this hypothesis.  All experiments are done on CheXpert for ResNet50, DenseNet121 and EfficientNet-b0.

The first experiment is a sanity check of our saliency metric.  We evaluate the effect of progressively zeroing out least (Fig. \ref{fig:sanity_check_saliency1}) and most (Fig. \ref{fig:sanity_check_saliency2}) salient spatial convolution filter kernels.  The results and extra details on experiment configuration are in Appendix \ref{appendix:sanity_check_saliency}.  We make two observations.  First, removing least salient spatial filters has no performance drop in some cases even when 50\% of kernels are zeroed.  Second, removing less than 3\% of the most salient filters makes the models have random output.  We conclude that (a) a very small portion of most salient spatial kernels are essential for good performance, (b) the majority of least salient kernels are unnecessary, and (c), most importantly, the saliency metric works.

Our next experiment verifies that few spatial filters are necessary for inference. 
Leveraging our explanation from Sec. \ref{results_explainability} that ImageNet pre-training sparsifies the ExplainSteer energy, we initialize 18 different networks with ImageNet Unchanged initialization, zero out the least salient filters, and fix all spatial layers.  We then independently train each network on CheXpert and compare it to a corresponding fully learned baseline.  We provide details in Appendix \ref{appendix:unnecessary} and results in Fig. \ref{fig:unnecessary_inference}.
We observe that 99\%, 95\% and 85\% of spatial filters are unnecessary for inference on CheXpert data in ResNet50, DenseNet121 and EfficientNet-b0, respectively.  Nearly all spatial filters are unnecessary for inference in spatially fixed models.

\subsubsection{Few Spatial Filters Suffice for Training}\label{sec:results_unnecessary_training}

In this experiment, we zero the fixed network before it is trained on CheXpert.  We start with 18 completely random (never trained) networks, fix all spatial weights and zero out the least salient filters.  Results and additional details are in Appendix \ref{appendix:unnecessary} and in Fig. \ref{fig:unnecessary_training}.  We observe that the fixed and zeroed models are nearly equal to the learned baseline and that 95\%, 90\% and 85\% of spatial filters are unnecessary for training non-spatial weights in ResNet50, DenseNet121 and EfficientNet-b0, respectively.  Most spatial filters are unnecessary for training non-spatial weights.

\subsubsection{Pruned Networks}\label{sec:results_pruning_subsection}
Our conclusions that most spatial filters are unnecessary for training imply that the several input and output channels across the network, including channels in non-spatial layers, can be removed from deep networks.  To verify this implication, we designed ChannelPrune to systemically remove input and output channels connected to the spatially zeroed kernels.  

Channel pruning removes channels from several connected layers.  The ImageNet pre-trained networks we prune have learned BatchNorm layers or bias terms that contribute predictive information regardless of whether the input activations have been zeroed at a spatial convolution layer.  It is thus possible that pruning these bias terms results in some performance loss, though we note three facts.  First, this bias can be approximately incorporated into the spatial layers by steering with the lowest frequency basis filter (an all ones matrix).  Second, the FillZero fixed initialization in Methods Sec. \ref{sec:pruning} is our solution to mitigate performance loss caused by this bias.  
Third, random networks that are fixed and spatially sparse (as done in Sec. \ref{sec:results_unnecessary_training}), guarantee exactly identical unchanged predictive performance before and after pruning because, in channels that are zeroed out, the corresponding ``ghost'' bias terms also remain zero during training (they cannot receive non-zero gradient updates).   

We conduct a sanity check on our pruning to verify that the pruned networks indeed have matching predictive performance to the fully learned baseline in Appendix \ref{appendix:comp_savings_same_acc}.  The results of the sanity check agree with our results in Sec. \ref{sec:results_unnecessary_inference} and Sec. \ref{sec:results_unnecessary_training}, and further reinforce our overall hypothesis the spatial weights of a deep network, even a pruned deep network, can be fixed.
Fixed and pruned \ExplainFix models, as shown in Fig. \ref{fig:compute_efficiency}, train faster than the fully learned baseline with matching accuracy.

\section{Discussion and Future Work}\label{discussion}

The current paradigm of deep learning is limited by its strong emphasis on weight optimization.  Weight optimization is over-emphasized.
\ExplainFix shows that most deep network weights can be removed before training.  All spatial convolution weights can be fixed.  Deep networks contain millions of weights, and yet most of these weights do not need to be learned or even used.  \ExplainFix is a significant step towards realization of fully fixed networks.  We next emphasize additional benefits of fixed weight deep networks in context of future research.

Standard deep networks define a fixed architecture at initialization.  The fixed architecture does not adapt to the input.  Architecture search is also computationally costly because the optimization occurs over both architecture and weights.
Fixed weight architectures eliminate optimization over weights. They enable dynamically generated or adaptive architectures and more efficient architecture search.  For instance, fixed weight networks can have essentially infinite size by dynamically expanding along selective paths, such as by the use of fixed weight attention mechanisms and a fixed weight operator such as a wavelet transform.  Fixed weight networks enable dynamic or learned architectures.

Towards explainability, development of fully fixed networks requires and encourages strong understanding of how deep networks work.  
An explanation should affect decision making.  Simply explaining predictions of single examples, for instance via interpretation of GradCam \parencite{gradcam} or Lime \parencite{lime}, may help a physician trust a model's output given a particular input, but it does not sufficiently explain the model itself or suggest ways to improve it.  The research literature needs \textit{model-based explanations} that lead to concrete suggestions on how to improve a model.  \ExplainFix offers such model-based explanations.  Our ExplainSteer visual explanations show that spatial weights are mostly unnecessary for training and inference especially in later layers and in higher frequency kernels.  We also explain that steered spatial filter initialization facilitating learning dynamics, and resulting in better performance.  \ExplainFix results in nimble, higher performing and better explained models.

Fully fixed deep networks would need no weight training.  They would enable development and democratization of larger and deeper networks at greatly reduced computational and data acquisition needs.  \ExplainFix's focus on spatial convolution filters marks a significant step towards fully fixed networks.  Future work should to consider fixing pointwise convolutions and ways to exploit trade-offs between fixed architectures and fixed weights.

\section{Conclusion}\label{conclusion}
\ExplainFix is adopts two design principles, the \fixedFilters principle that the weights of deep networks can be entirely fixed at initialization and never learned, and the \nimbleness principle that few parameters are necessary for training and inference.
We give three main contributions:
\begin{itemize}
  \item \textbf{Model-based Explanations:} We develop a saliency-based visualization tool to interpret spatial filter kernels in spatially fixed CNNS.  We show that fixed CNNS should have a steered representation to attain good performance, all spatial filters can be fixed, and most can be eliminated (up to 100x) from common deep network architectures on medical image data.
  \item \textbf{Speed and Accuracy Gains:} \ExplainFix models guarantee faster training (up to 17\% savings), smaller models with channel pruning (up to 5x fewer parameters), and they have matching or improved predictive performance.
Our novel steered initializations give winning ticket models that can even outperform fully learned networks.
\item \textbf{Novel Tools:} We contribute open sourced tools dedicated to fixed spatial filter networks, including three novel spatial filter initialization methods (GHaar, Psine, GuidedSteer), a novel explanation method (ExplainSteer), and a novel deep network pruning method (ChannelPrune).
\end{itemize}
\ExplainFix spans an extensive empirical analysis of four architectures, two medical image datasets, 13 baseline models and more than 330 distinct trained models.

\section*{Acknowledgments}
The project funding this work, Transparent Artificial Medical Intelligence (NORTE-01-0247-FEDER045905), is co-financed under the CMU-Portugal International Partnership by European Regional Fund through the North Portugal Regional Operational Program-NORTE 2020 (ERDF) and by the Portuguese Foundation for Science and Technology (FCT).

Special thanks to Shreshta Mohan for her review of the CheXpert loss function and correction of typographical errors.

\renewcommand*{\UrlFont}{\rmfamily}
\printbibliography

\title{Supplementary Appendices for ``\thepapertitle''}

\counterwithin{figure}{section}
\counterwithin{table}{section}

\appendix

  \section{Dataset Details}\label{appendix_dataset_details}

  \begin{figure}[H]
    \centering
      \centering
    \subfloat[][BBBC038v1 Dataset] {
      \centering
      \includegraphics[height=2.5cm]{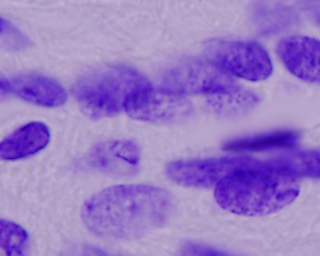}
      \includegraphics[height=2.5cm]{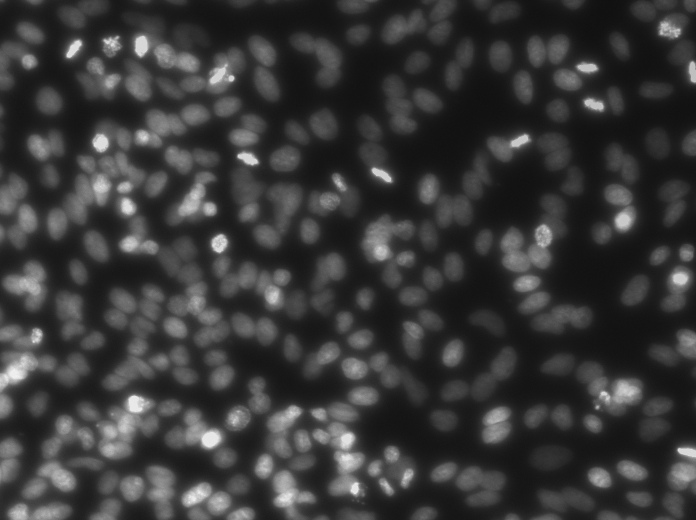}
      \includegraphics[height=2.5cm]{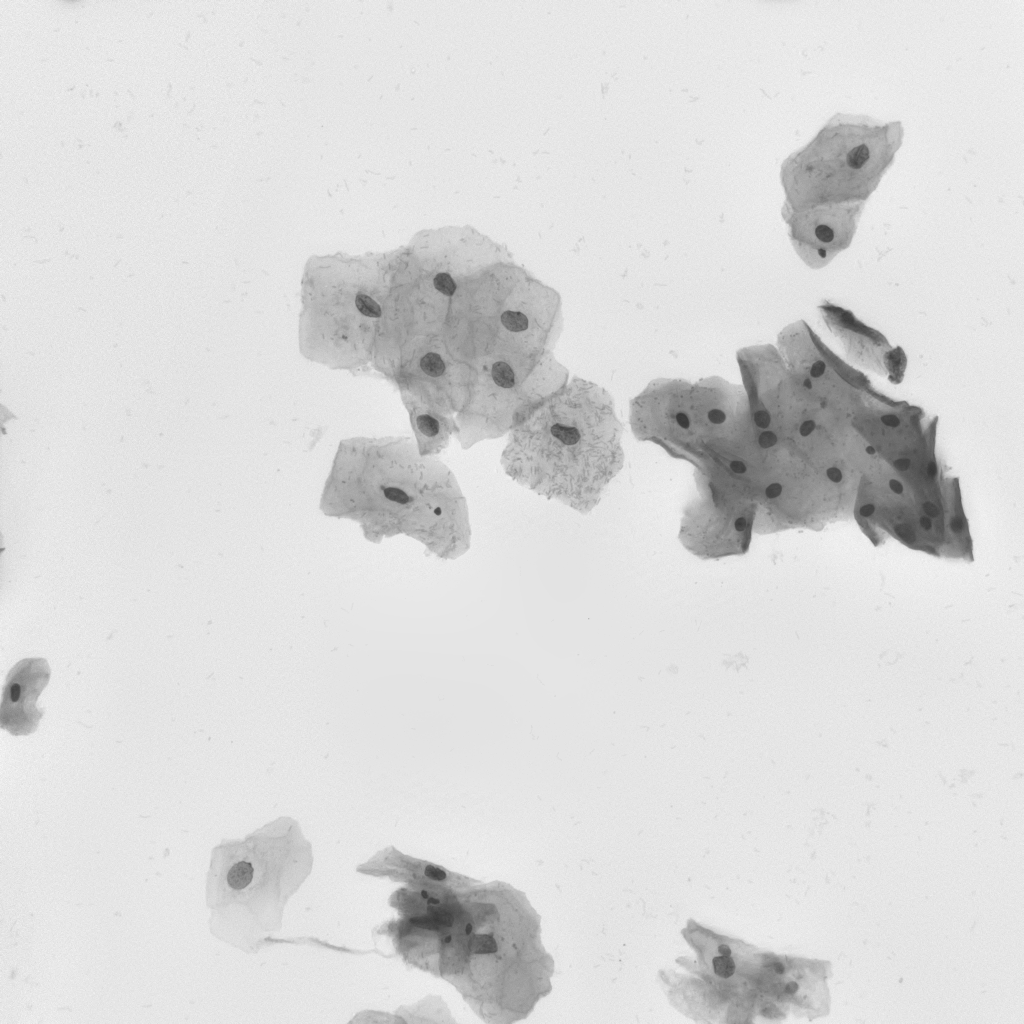}
      \label{fig:bbbc038v1_example}
    }
    \hfill
      \centering
    \subfloat[][CheXpert Dataset]{
      \centering
      \includegraphics[height=2.5cm]{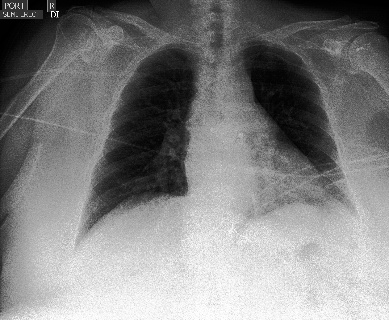}
      \includegraphics[height=2.5cm]{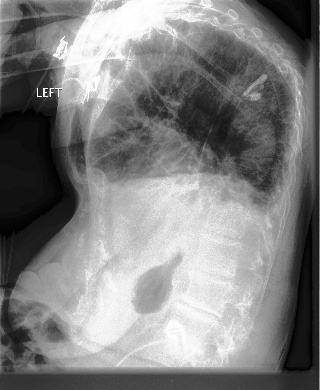}
      \label{fig:chexpert_example}
    }
    \caption{Example images from the datasets we used.}
  \end{figure}

  \textbf{CheXpert:}  The CheXpert dataset \cite{dataset_chexpert} published by Stanford ML Group and Stanford Hospital contains 223,415:235 train:test chest x-rays for binary classification of 14 classes.  Only five classes are emphasized by most literature, and we therefore consider just the five classes: Atelectasis, Cardiomegaly, Consolidation, Edema, and Pleural Effusion.  The train labels are generated by an automatic software that mines text data in associated unpublished medical reports, and the test set labels are manually curated by three radiologists.  Most labels are missing and some are marked uncertain.  There are several possible baselines that attempt to incorporate the uncertainty labels.  We compare against non-ensembled single-model U-Ignore baseline reported in the state of art literature \cite{chexpert_2019sota}, and we provide details on this baseline and comparison to state of art in Appendix \ref{appendix_model_details}.  This baseline assumes missing labels are negative and excludes uncertainty labels from the analysis.

  Regarding implementation, we use the CheXpert-v1.0-small dataset, which contains already down-sampled images.  We permanently split the training set into 70\%:30\% (156389:67025) using a random seed of 138 and PyTorch's default random number generator.  We train on the 70\% split and evaluate the 30\% validation set to find hyperparameters.  All results we report are on the 200 image holdout test set from models trained on our 70\% split of the training set.  For pre-processing, we randomly crop to images of $(320, 320)$ pixels on the 70\% training set, and on the validation or test sets we upsample images in a minibatch by padding zeros.  The minibatch size is four images.  Each epoch considers only 15k images sampled with replacement from the 70\% split.  To evaluate performance, we report the average test set ROC AUC across the five classes to enable comparison with most literature that uses the dataset and non-ensembled U-Ignore baseline.

\textbf{BBBC038v1:} The BBBC038v1 dataset \cite{dataset_BBBC038v1}, hosted by the Broad Bioimage Benchmark Collection, is a diverse collection of microscopy images visualizing cell nuclei.  The dataset is representative of the domain: images are aggregated from a variety of laboratories, they show cell nuclei of different animals in different states and contexts, and present varying lighting, magnification, staining, and background/foreground coloration.  The dataset contains 670:65:106 images annotated with pixel-wise segmentation masks.  The 106 hold-out images are out-of-distribution, as they come from entirely different laboratories and they present experimental conditions not in the 670:65 data.  We treat both the 65 and 106 image datasets as test sets in order to have an in-distribution vs out-of-distribution analysis.  We visualize example images in Fig. \ref{fig:bbbc038v1_example}.

To choose hyperparameters, data pre-processing and loss, we temporarily split the training set into 468:202 training and cross validation sets using a different random seed each time we train a model.  We report all results on the 106 image holdout set, except in Appendix \ref{appendix:BE11_in_vs_out_of_distribution} where we show results on the in-distribution vs out-of-distribution datasets.  
Regarding pre-processing, all images were normalized into [0,1] and then shifted to have a 0.5 mean.  On the training set, the images were subjected to random flipping (even probability of no flip, horizontal, vertical, both), and then with 80\% probability a pixel-wise noise function for each pixel $x' = \text{\texttt{clip}}(x + \mathcal{N}(0,1) / \mathcal{U}(5,10))$ where \texttt{clip} puts the image into a [0,1] range.  We use a batch size of ten on the training set, and one on the validation and test sets.  We report results using the S{\o}rensen-Dice Coefficient evaluated on the test set.

  \section{Model Details} \label{appendix_model_details}

  We consider three standard architectures for the CheXpert dataset: DenseNet \cite{densenet}, ResNet \cite{resnet} and {EfficientNet} \cite{efficientnet}.  DenseNet121 is used by state-of-the-art literature on chest x-ray data, including CheXpert \cite{chexpert_2019sota}.  ResNet introduced the residual skip connection and the bottleneck sequence of convolutions $1\times1$, $3\times3$, $1\times1$. Sec. \ref{intuitions} explains that this sequence enables the network architecture to perform steering of inputs and outputs to the spatial convolution.  The DenseNet architecture builds on ResNet's skip connection with "dense" blocks and is largely defined as a sequence of $1\times 1$, $3\times3$ convolutions, thus steering spatial convolution inputs.
  The EfficientNet architecture uses the depthwise separable convolution with both $3\times3$ and $5\times5$ filters, though the network primarily contains of pointwise convolutions.  Table \ref{table:model_num_params_and_savings} (middle column) compares the number and proportion of spatial parameters in each model.  The pre-trained models use publicly available ImageNet weights, and for pre-trained EfficientNet we use the weights obtained via the AdvProp method.  All models are modified to have 1 channel grayscale image input and output an unnormalized vector with five values corresponding to the five CheXpert tasks.

  \begin{figure}[t]
    \centering
    \subfloat[CheXpert Baselines\label{fig:chexpert_baselines}]{
      \includegraphics[height=3.5cm]{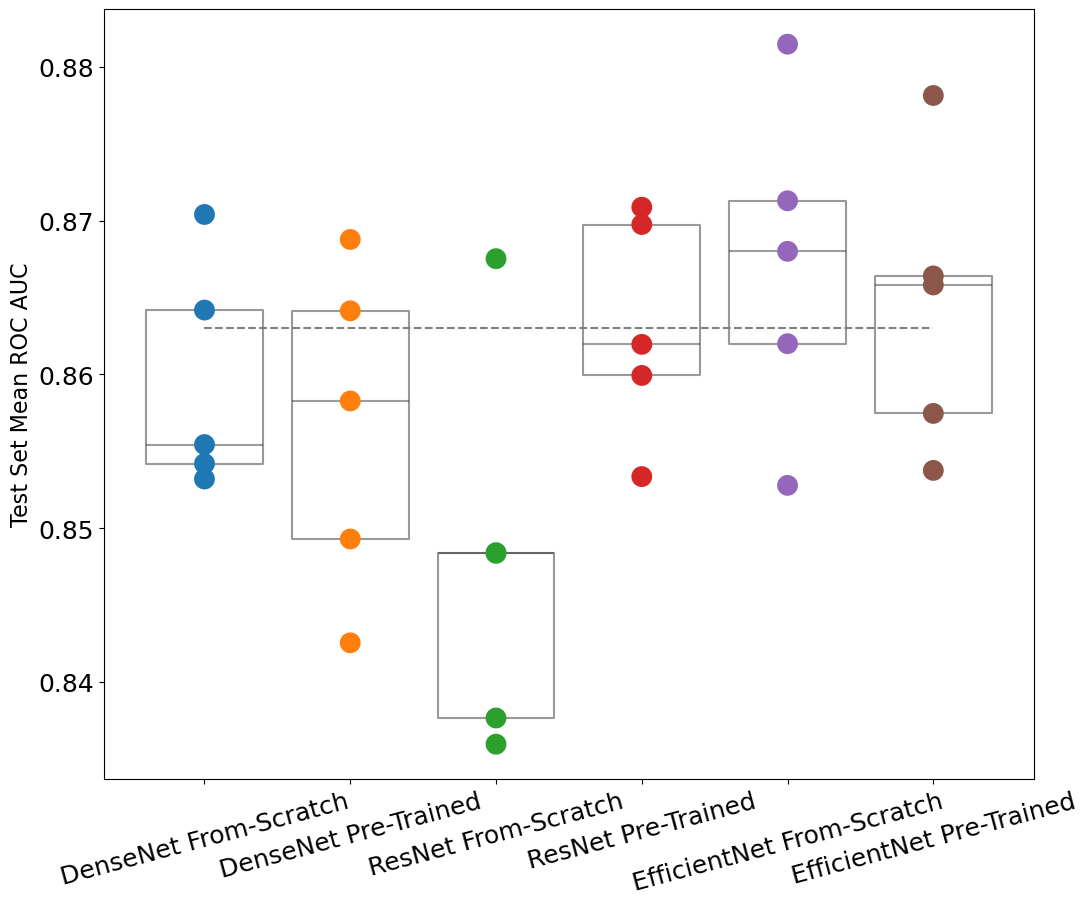}
    }\qquad
    \subfloat[BBBC038v1 Baselines\label{fig:bbbc038v1_baselines}]{
    \includegraphics[height=3.5cm]{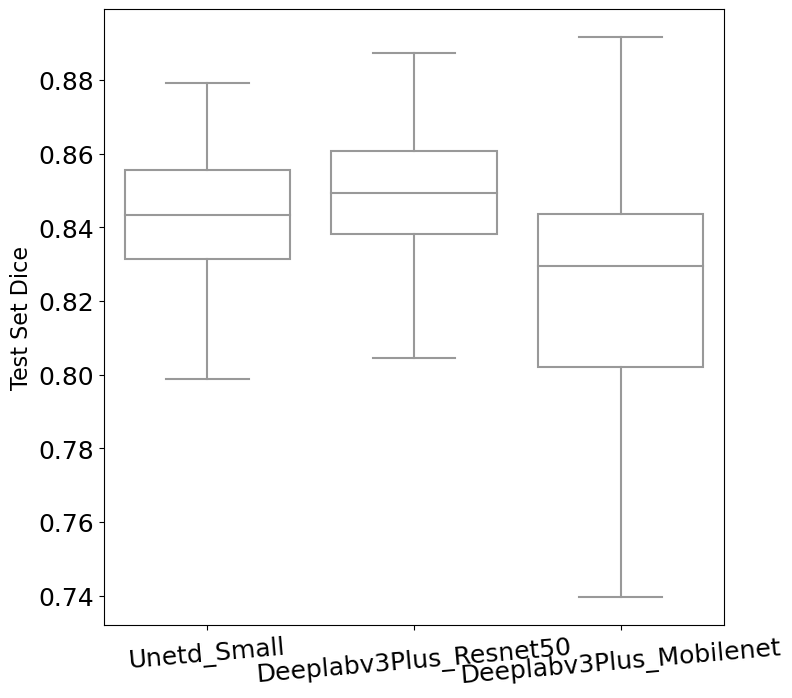}
    }\qquad
    \subfloat[U-NetD Architecture (simplified) \label{fig:unetD}]{
    \includegraphics[height=3.5cm]{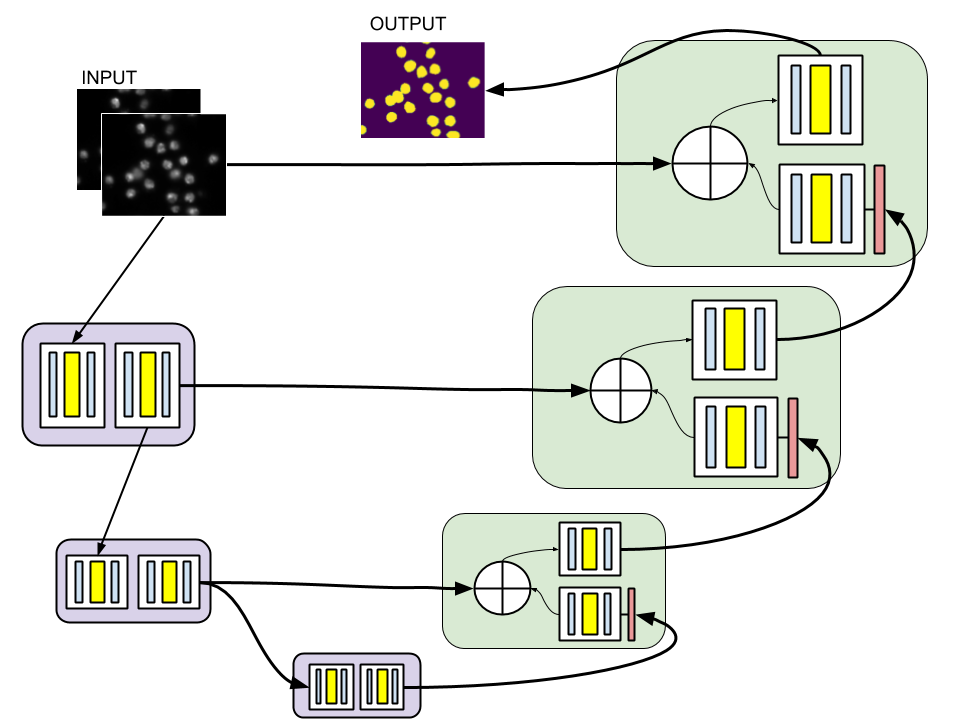}
    }
  \end{figure}
  \begin{table}[t]
    \centering
    \caption{BBBC038v1 Baselines: U-NetD has fewest parameters}
    \label{table:bbbc_num_params}
    \begin{tabular}{lrr}
    \toprule
    {Model} &  Num Params & \begin{tabular}{@{}c@{}}Spatial Params\\(\% total)\end{tabular} \\
    \midrule
    U-NetD             &      116695 &             12744 (11\%) \\
    DeepLabV3+ MobileNetV2 &     5220577 &           2977344 (57\%) \\
    DeepLabV3+ ResNet50  &    39756705 &          26182848 (66\%) \\
    \bottomrule
    \end{tabular}
  \end{table}
  On the CheXpert U-Ignore baseline, current state-of-the-art literature reports a mean AUC ROC of 0.863, using DenseNet121 with a full training set \cite{chexpert_2019sota}, meaning that each training image is considered five times in total (five epochs).  
All our baseline models are comparable to this baseline (see Fig. \ref{fig:chexpert_baselines}), and our models
use significantly less training data (150k training images instead of 223k, and each image is used for training only four times on average).  In Fig. \ref{fig:chexpert_baselines}, each dot represents an independent training of one of our baseline models.  The gray dashed line is the 0.863 ROC AUC from existing literature.

  On the microscopy dataset, we implement our own depthwise-separable {U-Net} \cite{unet_ronneberger2015} encoder-decoder architecture, called U-NetD.  For clarity, the architecture is visualized with only 3 levels in Fig. \ref{fig:unetD}. We use 5 levels, each with a different channel width: $(3,8,16,32,64)$.  In the figure, green blocks are spatial $3\times3$ grouped convolutions with 6x channel expansion, blue blocks are pointwise convolutions, and red blocks are bilinear upsampling.  Between convolutions, we insert CELU activations and then BatchNorm, though the pointwise convolution following every spatial convolution has no activation, as suggested in MobileNetV2 \cite{mobilenetv2}.  We fuse the encoder output with previous decoder output using a weighted sum with two learned scalar weights.  
  The final layer of U-NetD (not shown in figure) is a 2D spatial convolution that maps its 3 channel output to a 1 channel segmentation mask.  All spatial convolutions are $3\times3$ kernels with no bias.

  We compare this U-Net model against DeepLabv3+ with the ResNet50 and MobileNetV2 backbones in Fig. \ref{fig:bbbc038v1_baselines}.  Each architecture presented was trained three times independently and evaluated on the out-of-distribution BBBC038v1 test set.  The boxenplot shows test set segmentation performance after 150 epochs.  The U-NetD model outperforms the MobileNetV2 backbone.  It converges almost the same median value as the ResNet50 backbone.  U-NetD has one and two orders of magnitude fewer parameters than the MobileNet and ResNet50 DeepLabV3+ architectures, respectively (see Table \ref{table:bbbc_num_params}).  We adopt the U-NetD over the DeepLabv3+ models because (a) the ResNet backbone is already considered in the CheXpert models, (b) U-NetD follows MobileNetV2 recommendations in its design (MobileNetv2 is desirable to emulate for its steering and wavelet-like properties as described in Sec. \ref{intuitions}), (c) U-NetD presents another common kind of architecture (the U-Net), and (c) U-NetD has orders of magnitude fewer parameters.

  \section{Other Hyper-parameters and Loss Functions} \label{appendix_hyperparameter_details}

  \textbf{Hyper-parameters:}  Dataset and model-specific configuration is defined in Appendices \ref{appendix_dataset_details} and \ref{appendix_model_details}.  In fixed filter models, all spatial filter kernels are not modified by back propagation.  Unless explicitly stated, we train all CheXpert models for 40 epochs using the default Adam optimizer with learning rate 0.0001, and we train all BBBC038v1 models for 300 epochs using the default Adam optimizer with learning rate 0.008.  Pruned models have a 2x larger learning rate of 0.16.

  The BBBC038v1 Loss function is a pixel-wise binary cross entropy loss with no class balancing weights.

  \textbf{CheXpert Loss function:} We designed a multi-label focal binary cross entropy loss with class balancing weights. The unreduced loss is defined as a $4\times5$ matrix:
\begin{align}
  L = - \mathbf{m} \w_{\text{task}} \bigl( \w_{\text{pos}}\y \log (\hat \y) + \w_{\text{neg}}(1-\y) \log (1- \hat \y) \bigr),
\end{align}
where $\y\in[0,1]^{4\times5}$ is ground truth for each of the five classes across the minibatch of 4 images, $\hat \y \in [0,1]^{4\times5}$ is the model output with each value normalized by a rescaled sigmoid function  $\hat \y = 0.99999\sigma(\hat \y') + 0.000005)$.  The matrix $\mathbf{m}\in\{0,1\}^{4\times5}$ is a bitmask that ignores uncertainty labels, the multiplications are element-wise multiplication with broadcasting, and the class inter- and intra-class balancing weights: $\w_\text{task} \in \mathbb{R}^{1\times5}$ balances the total count of positive + negative labels across tasks; $\w_\text{pos}\in\mathbb{R}^{1\times5}$ and $\w_\text{neg}\in\mathbb{R}^{1\times5}$ balance the counts of positive and negative classes within any given class (specifically ignoring uncertainty labels and remapping missing labels to negative) and incorporate a focal term.  For each of the classes, we have scalar weights:

  \begin{minipage}{.48\linewidth}
  \begin{align*}
    w_\text{pos} &= a_\text{pos}(1-\hat y)^\gamma \\
    \\
    w_\text{neg} &= a_\text{neg}\hat y^\gamma
  \end{align*}
  \end{minipage}%
  \begin{minipage}{.48\linewidth}
  \begin{align}
    a_\text{pos} &= \frac{1}{1 + \frac{c_\text{pos}}{c_\text{neg}}} \\
    a_\text{neg} &= \frac{1}{1 + \frac{c_\text{neg}}{c_\text{pos}}}
  \end{align}
  \end{minipage}
  where $\gamma=1$ is the focal loss hyperparameter that places emphasis on "hard" samples the model has low confidence for, where $a_\text{pos}$ and $a_\text{neg}$ are obtained by using standard class balancing weights of the vectorized form $\x = \frac{\max(\mathbf{c})}{\mathbf{c}}$ and subsequently normalizing them using $f(\x) = \frac{\x}{\sum_i x_i}$. The values $\mathbf{c} = [c_\text{pos}, c_\text{neg}]$ are counts of samples with positive and negative labels respectively over the entire training dataset for the given class.  For the inter-class balancing weight, we have a similar definition:
  \begin{align}
    \mathbf{c} &= \begin{bmatrix} c_{t} \;\;\forall\; \text{tasks }t \;\;;\;\; c_{t} = (c_{\text{pos}}+c_{\text{neg}})_{(t)}\end{bmatrix}\\
    w_\text{task} = w_t &= \frac{1}{1 + c_t \sum_{i\ne t} \frac{1}{c_i}}
  \end{align}
  We aggregate the loss $L$ by summing across columns (tasks) and then computing a mean across rows (minibatch samples). 

\section{Out-of-Distribution vs In-Distribution Analysis on Fixed Filters}\label{appendix:BE11_in_vs_out_of_distribution}

\begin{figure}[H]
  \centering
  \includegraphics[width=0.49\linewidth]{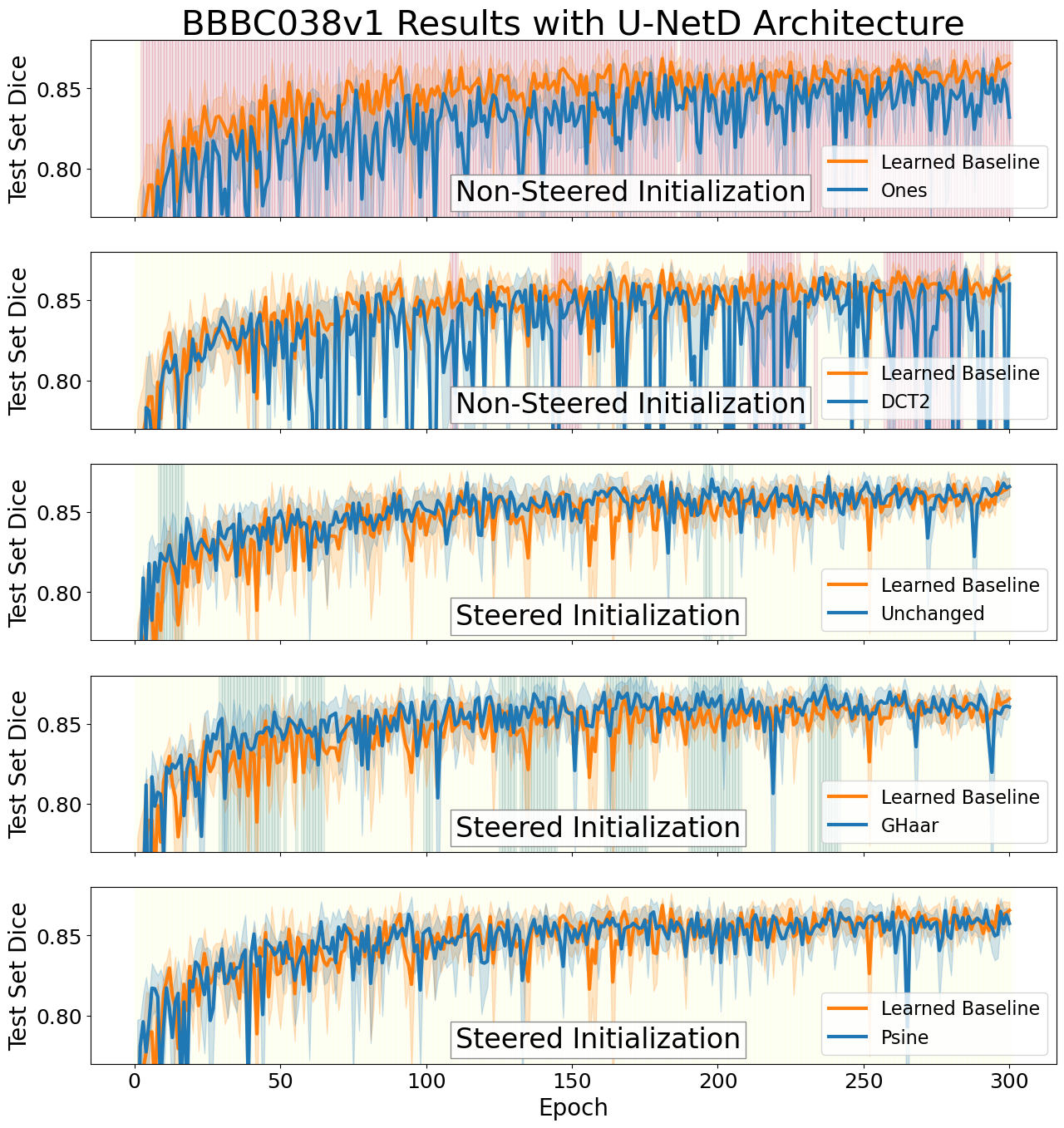}
  \includegraphics[width=0.49\linewidth]{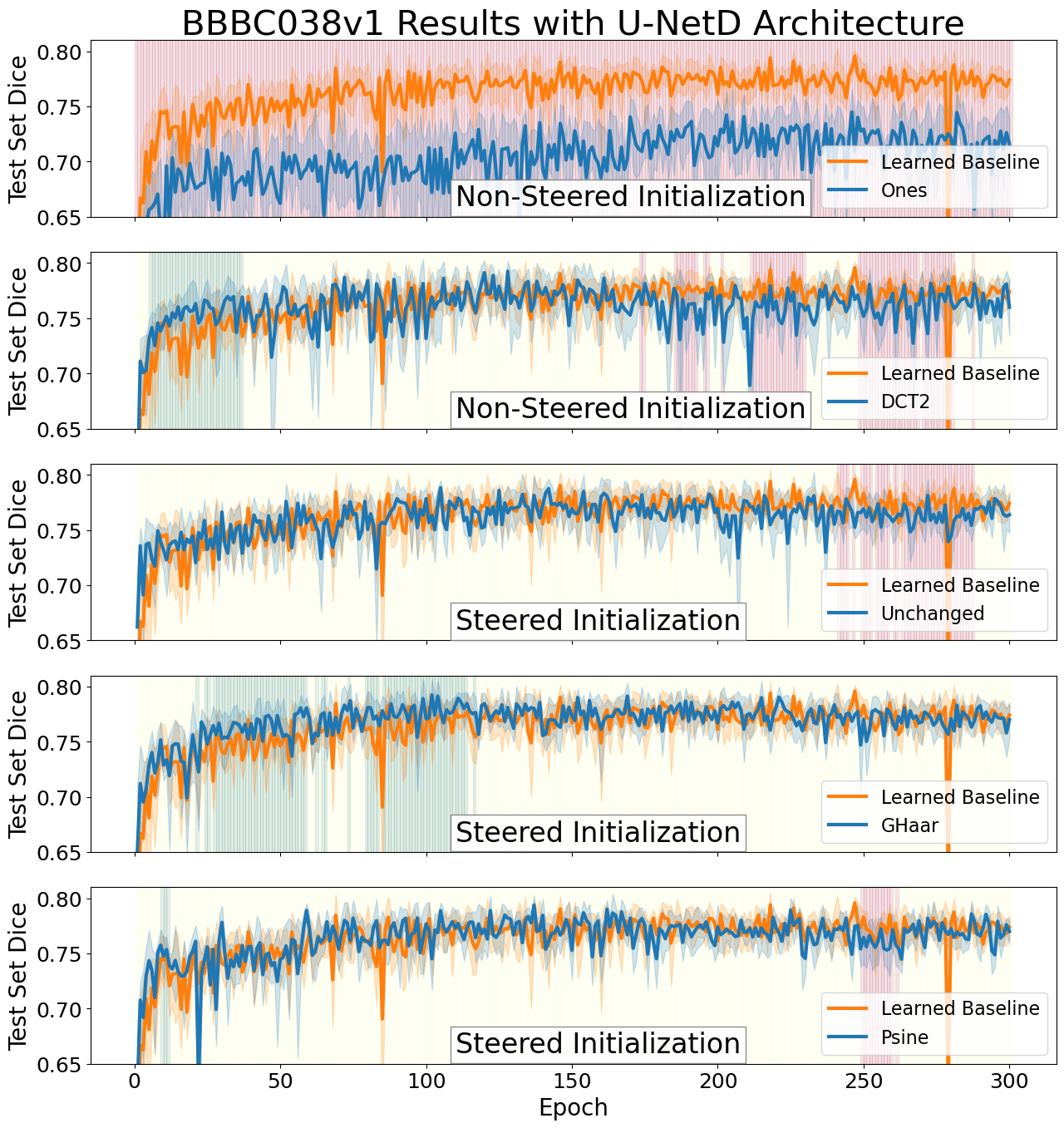}
  \caption{\textbf{No Overfitting.}
Left: Microscopy results of 36 independently trained models on the out-of-distribution test set.  Right: Otherwise identical experiment, on the in-distribution test set (same as Fig. \ref{appendix:BE10_results} left).  We observe higher performance on the out-of-distribution dataset.
\\
Each sub-figure represents 36 independently trained models.  Blue and orange lines show the per-epoch average of the same fixed spatial filter model (varying only in initialization) trained six times independently, and the blue line is the same in all plots.  The red and green background correspond to one-tailed paired-sample Wilcoxon significance tests \cite{wilcoxon_test1945}, one for each epoch, of whether the fixed filter models tended to outperform ($p>.999$) or underperform ($p<.001$) the baseline across all initializations.  Each test evaluated six paired models in a bandwidth of $\pm10$ epochs and across six initializations.  
}
\end{figure}

\section{Robustness to Learning Rate on Microscopy Dataset} \label{appendix:BE10_results}
\begin{figure}[H]
  \centering
  \includegraphics[width=0.44\linewidth]{pics/e1-BE11-results.png}
  \qquad
  \includegraphics[width=0.44\linewidth]{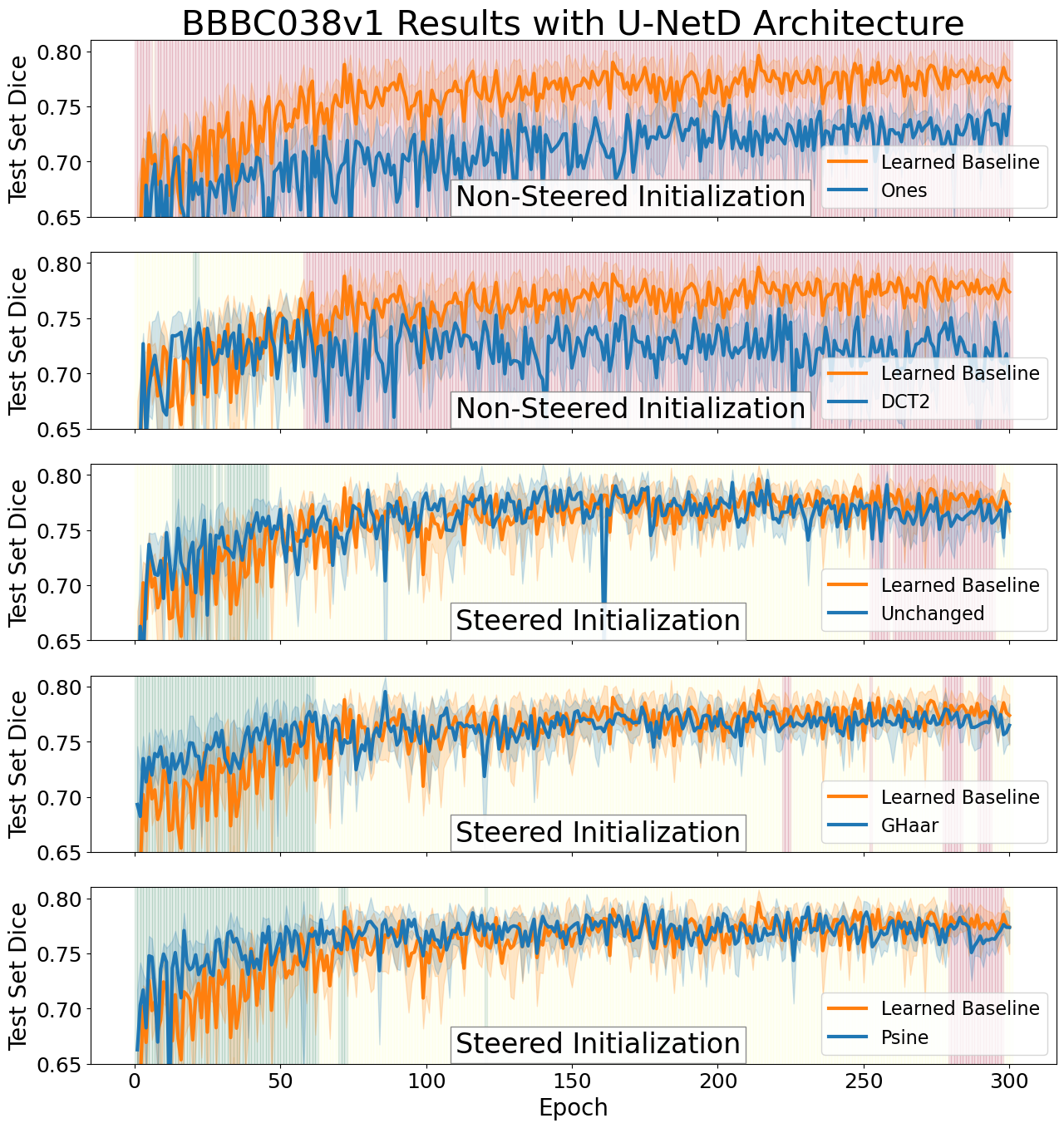}
  \caption{\textbf{Robust to increased Learning Rate.}  Left: Showing microscopy results of 36 independently trained models and learning rate .008.  Right: Otherwise identical experiment, with learning rate .02.  With larger learning rate, steered methods outperform the baseline early in training, and the non-steered DCT2 method underperforms.  The results suggests steered initialization improves robustness of the model.  Evaluated on BBBC038v1 in-distribution test set.}
\end{figure}

  \section{DCT-II Basis} \label{appendix_DCT-II}
\begin{figure}[H]
  \begin{center}
    \subfloat[\label{fig:DCTII_bases3}]{ \includegraphics[width=.3\linewidth]{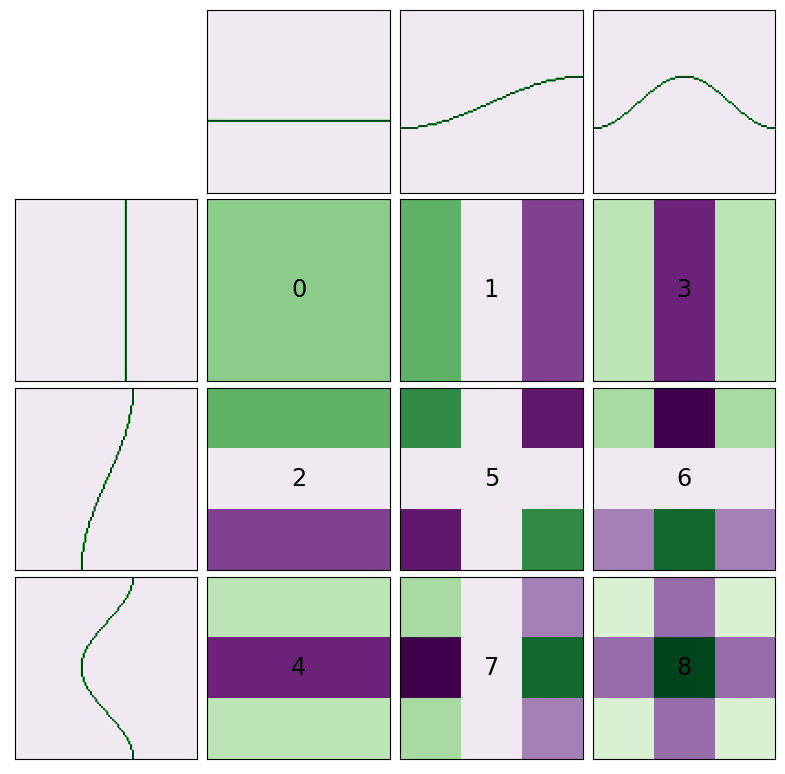}}
    \qquad
    \subfloat[\label{fig:DCTII_bases5}]{ \includegraphics[width=.3\linewidth]{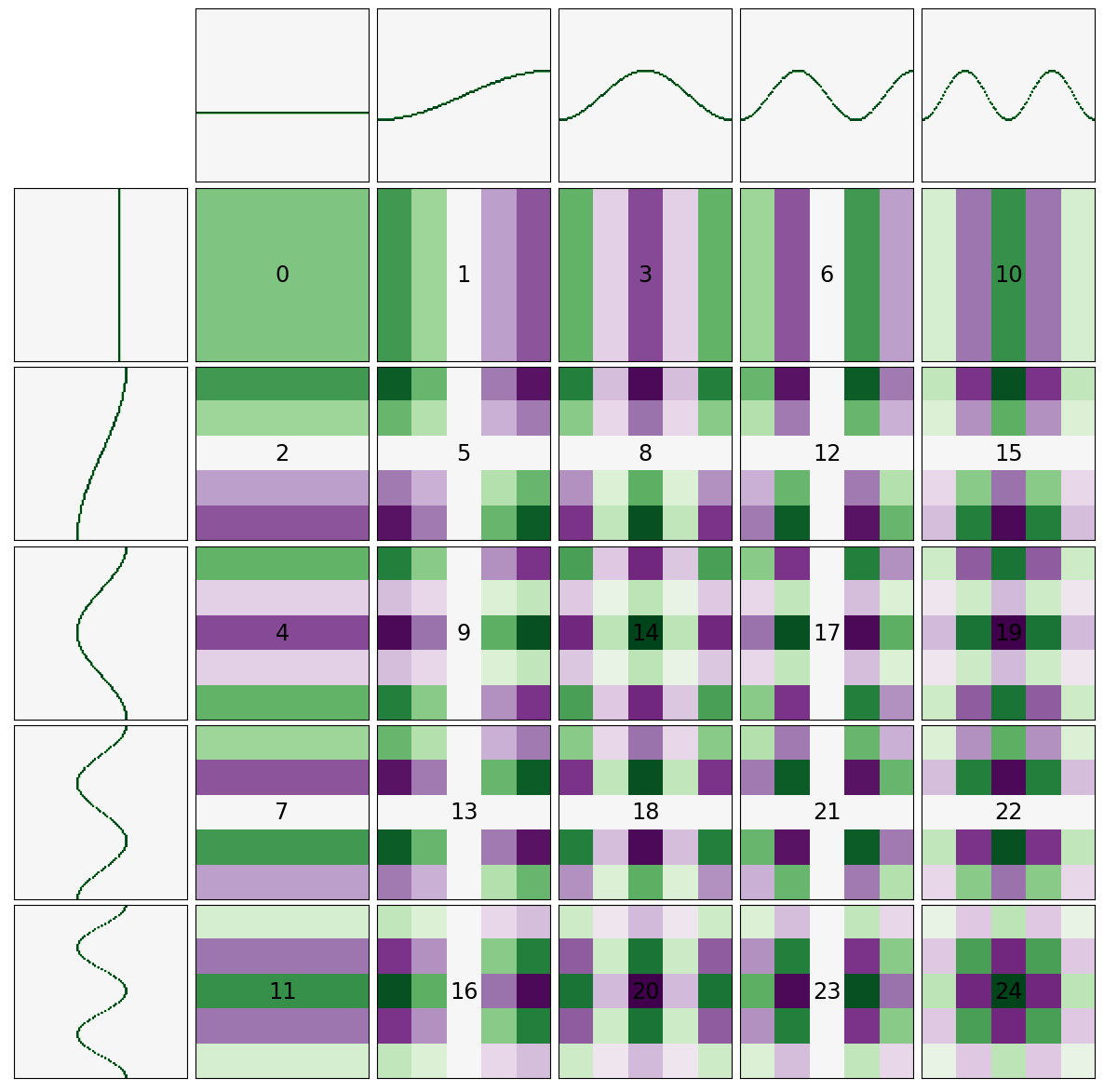}}
  \end{center}
  \caption{\textbf{Illustration of proposed basis} for $3\times3$ (a) and $5\times5$ (b) matrices. The index numbers order filters from lowest frequency basis filters to highest frequency, in stable ordering.}
  \label{fig:DCTII_bases}
  \end{figure}

  We visualize the 2-d DCT-II basis for $3\times3$ and $5\times5$ kernels in Fig. \ref{fig:DCTII_bases}.  The 2-d DCT-II is 1-d separable, and the first row and column show which the 1-d basis vectors construct each 2-d vector.  We also assign each basis filter an index value to sort filters from low to high frequency.

\section{ExplainSteer Explanations, Across Architectures and Initializations} \label{appendix:saliency_vs_not}

\begin{figure}[H]
  \centering
    \subfloat[]{\includegraphics[height=1.5cm]{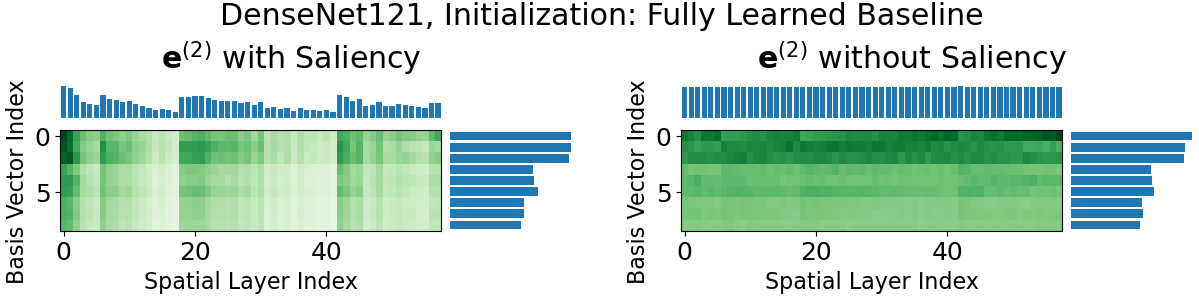}}
    \subfloat[]{\includegraphics[height=1.9cm]{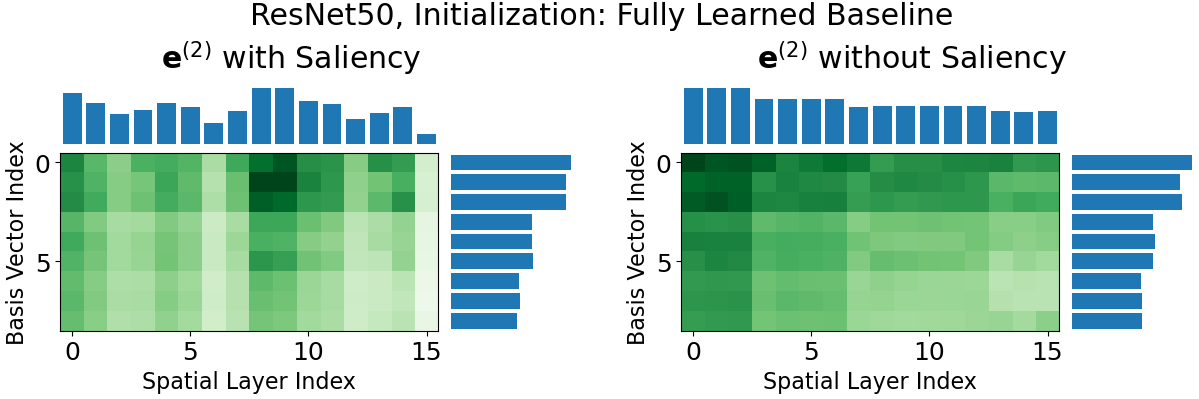}}
    \subfloat[]{\includegraphics[height=2.5cm]{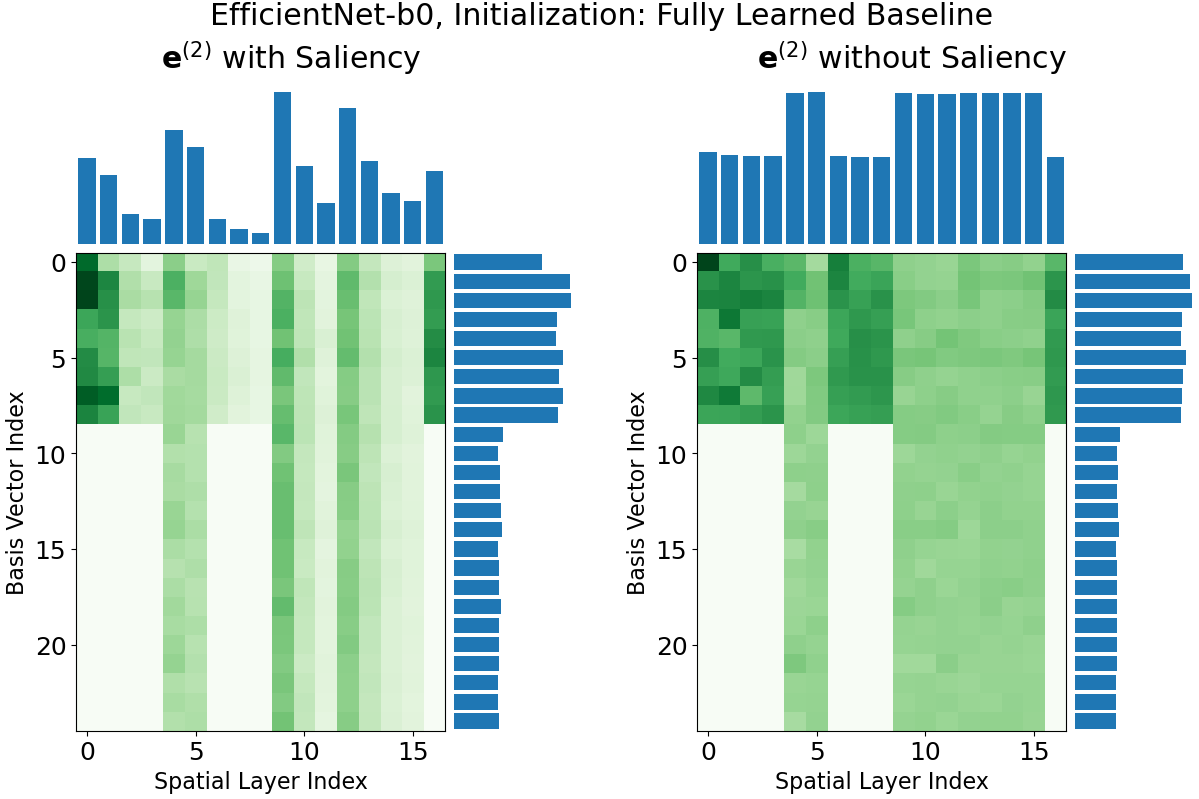}}
  \caption{\small \textbf{Illustration of Nimbleness} across architectures.  Saliency heatmaps expose structural inefficiency in the network design, where the repeating horizontal trend from bright to dark in all three architectures suggests that later spatial layers of most stages in the networks are less useful to the network.  All plots represent a fully learned baseline model initialized with random weights and trained on the CheXpert dataset.}
  \label{fig:explainsteer_across_architectures}
\end{figure}

\begin{figure}[H]
  \centering
  \subfloat[]{ \includegraphics[width=0.40\linewidth]{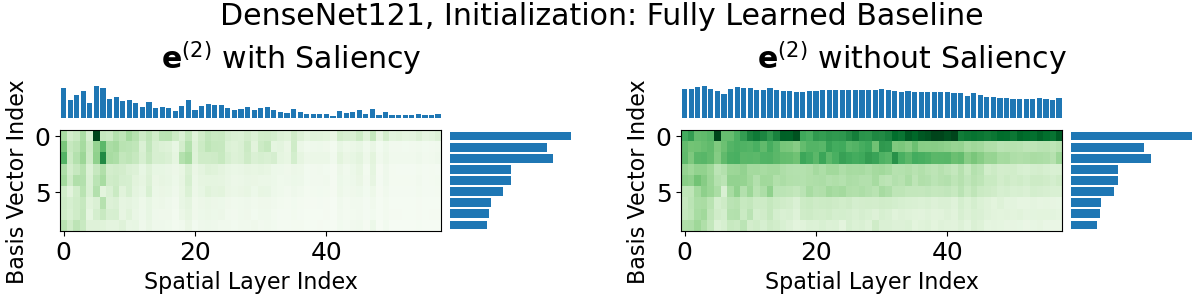}}
  \subfloat[]{ \includegraphics[width=0.40\linewidth]{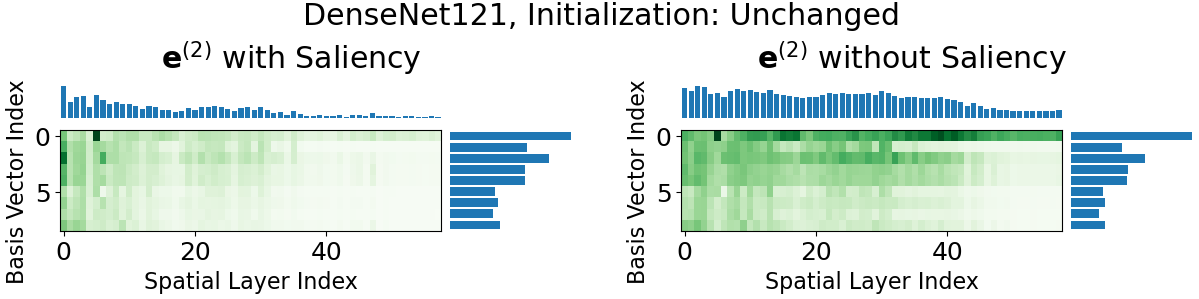}}\\
  \subfloat[]{ \includegraphics[width=0.40\linewidth]{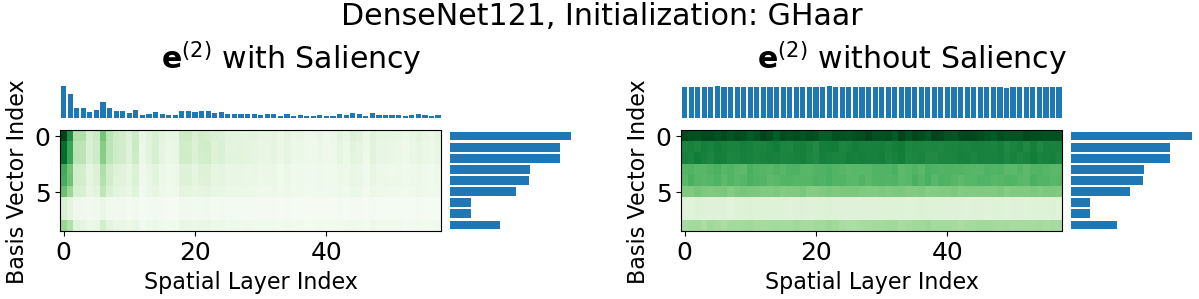}}
  \subfloat[]{ \includegraphics[width=0.40\linewidth]{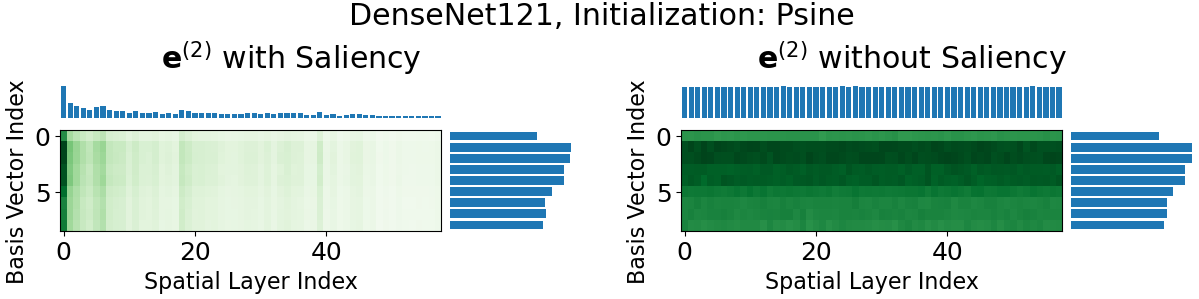}}\\
  \subfloat[]{ \includegraphics[width=0.40\linewidth]{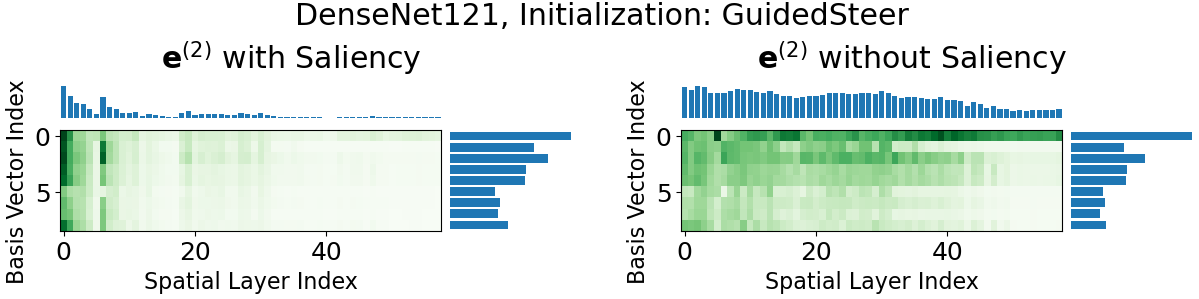}}
  \subfloat[]{ \includegraphics[width=0.40\linewidth]{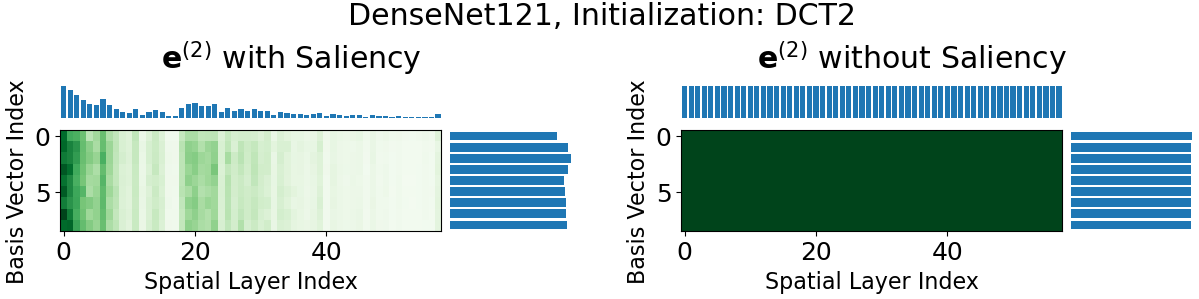}}
  \caption{\small \textbf{Illustration of Nimbleness} across fixed initializations.  Both horizontal and vertical trends reveal redundancy and inefficiency built into DenseNet121.  Each sub-figure is a DenseNet121 model pre-trained on ImageNet, spatially re-initialized with one of our fixed spatial filter methods, and fine-tuned on the CheXpert dataset.  All saliency heatmaps show that later layers are less influential (darker from left to right), and that higher frequency basis filters are less influential (darker from top to bottom).  This visual explanation motivates pruning the network.  
    \\\\
  \textit{Sanity checks:}  The GuidedSteer and Unchanged methods should appear nearly identical in "without Saliency" heatmaps.  The DCT2 heatmap without saliency should evenly distribute energy across basis vectors, resulting in a flat green heatmap.  GHaar and Psine should prioritize lower frequencies.  All sanity checks pass.
}
  \label{fig:explainsteer_across_initialization_methods}
\end{figure}

We visualize the $\e^{(d)}$ spectra for the spatial convolution layers of three architectures (DenseNet121, ResNet50, EfficientNet-b0), with varying initialization methods.  
The EfficientNet models have $3\times 3$ and $5\times 5$ filters and we visualize both kernel shapes in the same heatmap.  In DenseNet models, we do not visualize the spectrum of first convolution layer because this layer, which uses a 7x7 kernel, would result in a single column with 49 rows while the remaining columns all have 9 rows.  Below, Fig. \ref{fig:explainsteer_across_architectures} compares across architectures and Fig. \ref{fig:explainsteer_across_initialization_methods} compares across initialization methods.


\section{ResNet50 Model:  A Winning Ticket on the 67k Image CheXpert Validation Set}
\label{appendix_effects_of_pretraining}

\begin{figure}[H]
  \centering
  \includegraphics[width=0.6\linewidth]{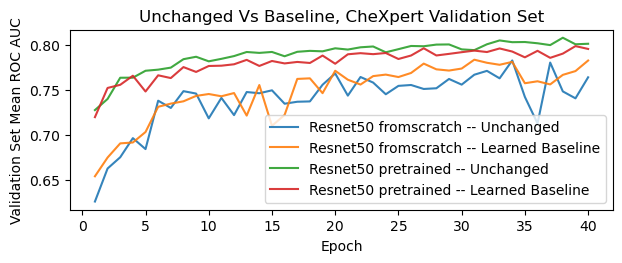}
  \caption{\textbf{Illustration of a Winning Ticket Model.} On ResNet50, the pre-trained Unchanged initialization outperforms the fully learned pre-trained baseline while random Unchanged initialization underperforms.  This finding is also reflected in test set performance in Fig. \ref{fig:chexpert_C8}, suggesting the pre-trained model has a winning ticket initialization.  The validation set has 67k images and is generally representative of test set performance.}
\end{figure}

  \section{Sanity Check on Spatial Filter Saliency}
  \label{appendix:sanity_check_saliency}
  \begin{figure}[H]
    \centering
    \subfloat[\label{fig:sanity_check_saliency1}]{\includegraphics[width=0.45\linewidth]{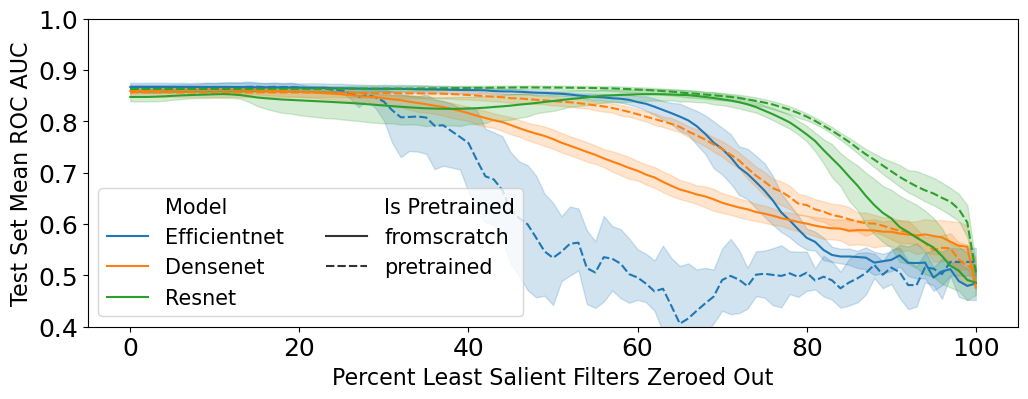}}
    \subfloat[\label{fig:sanity_check_saliency2}]{\includegraphics[width=0.45\linewidth]{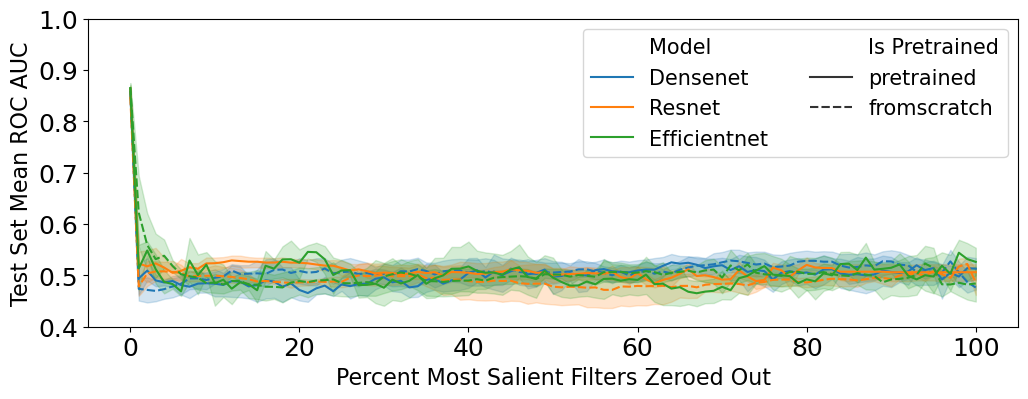}}\\
    \subfloat[]{\includegraphics[width=0.45\linewidth]{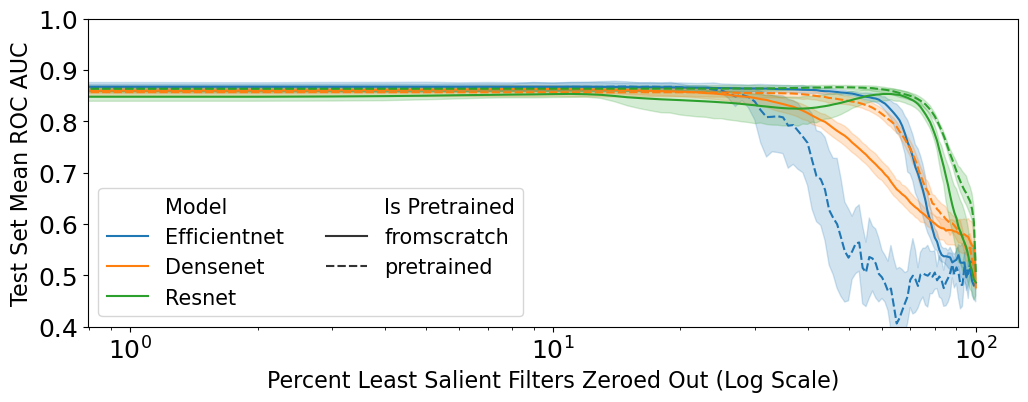}}
    \subfloat[]{\includegraphics[width=0.45\linewidth]{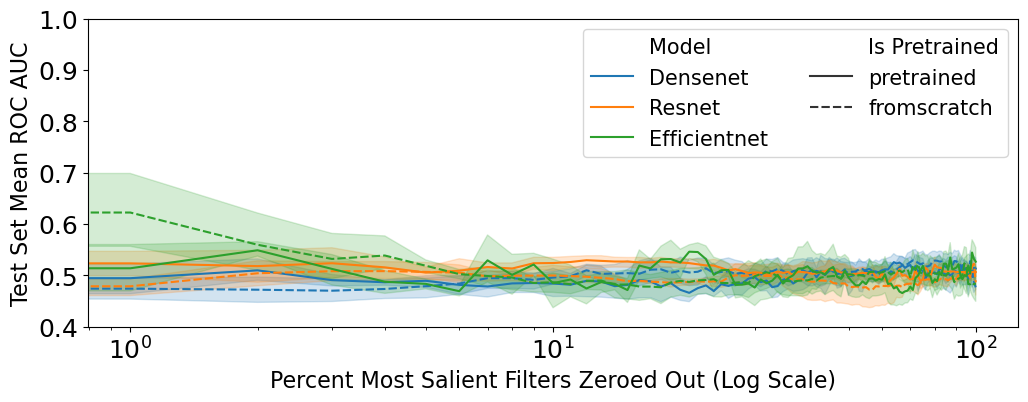}}
    \caption{\textbf{Proposed Saliency Method Works Well.}  This sanity check experiment shows the effect of progressively replacing $x$ percent of most salient (left) and least salient (right) spatial filter kernels with zeros.  Bottom row: x-axis on log scale.  In plots, each line represents six independently trained Learned Baseline models with a 95\% confidence interval.  The models were first trained on CheXpert, fixed and then evaluated while spatial kernels were progressively zeroed using our saliency metric in Eq. \ref{eq:saliency}.  
    The plots on right are a sanity check that the most salient filters are actually necessary for inference.  Removing most important filters indeed makes the models random.  All plots suggest that few spatial filters are relevant to the prediction task.  All plots confirm that our saliency metric separates important from unimportant spatial filters.}
  \end{figure}

\section{Visualizing Unchanged Filters}\label{sec:appendix_Unchanged}

\begin{figure}[H]
    \centering
    \subfloat[Random Unchanged]{\includegraphics[width=0.48\textwidth]{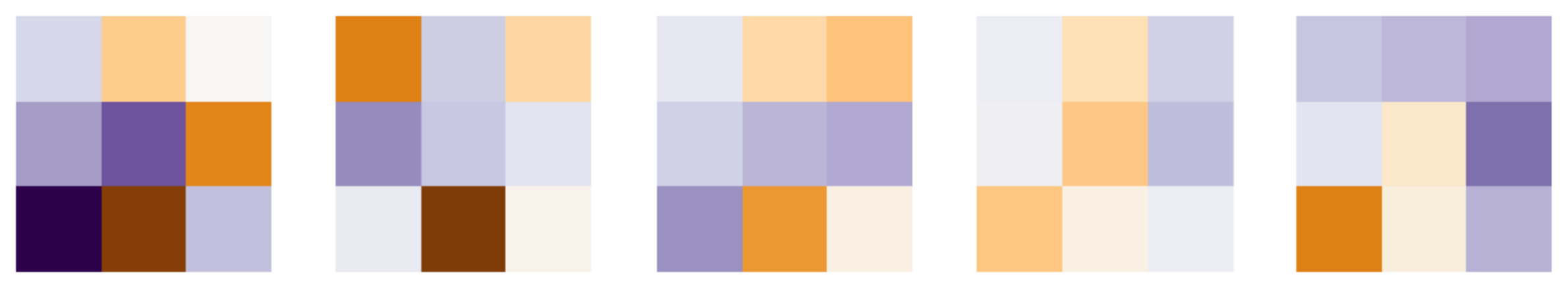}}\hfill
    \subfloat[ImageNet Unchanged]{\includegraphics[width=0.48\textwidth]{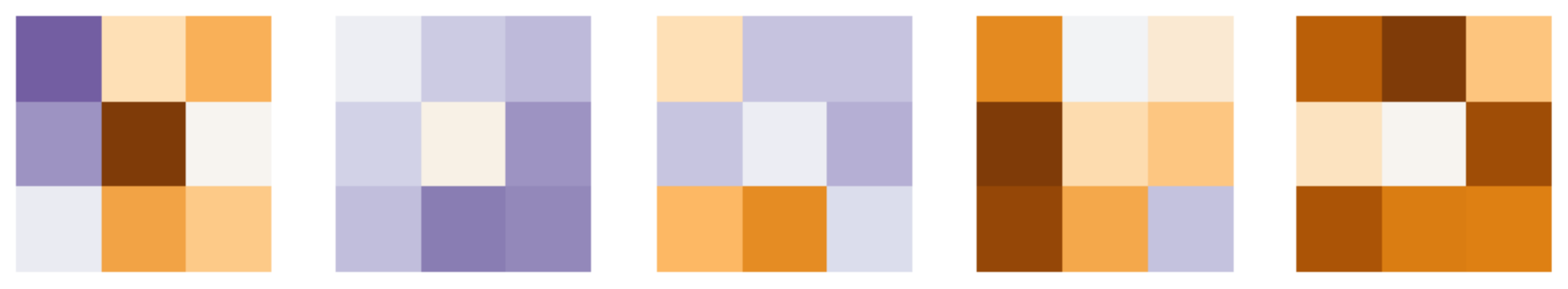}}
    \caption{\textbf{Visualization of Unchanged filters} for a DenseNet121 model at the first 3$\times$3 convolution layer.}
\end{figure}

\section{Most Spatial Weights are Unnecessary for Inference and Training}\label{appendix:unnecessary}
  \begin{figure}[H]
    \centering
    \subfloat[Unnecessary for Training\label{fig:unnecessary_training}]{
    \includegraphics[width=.49\linewidth]{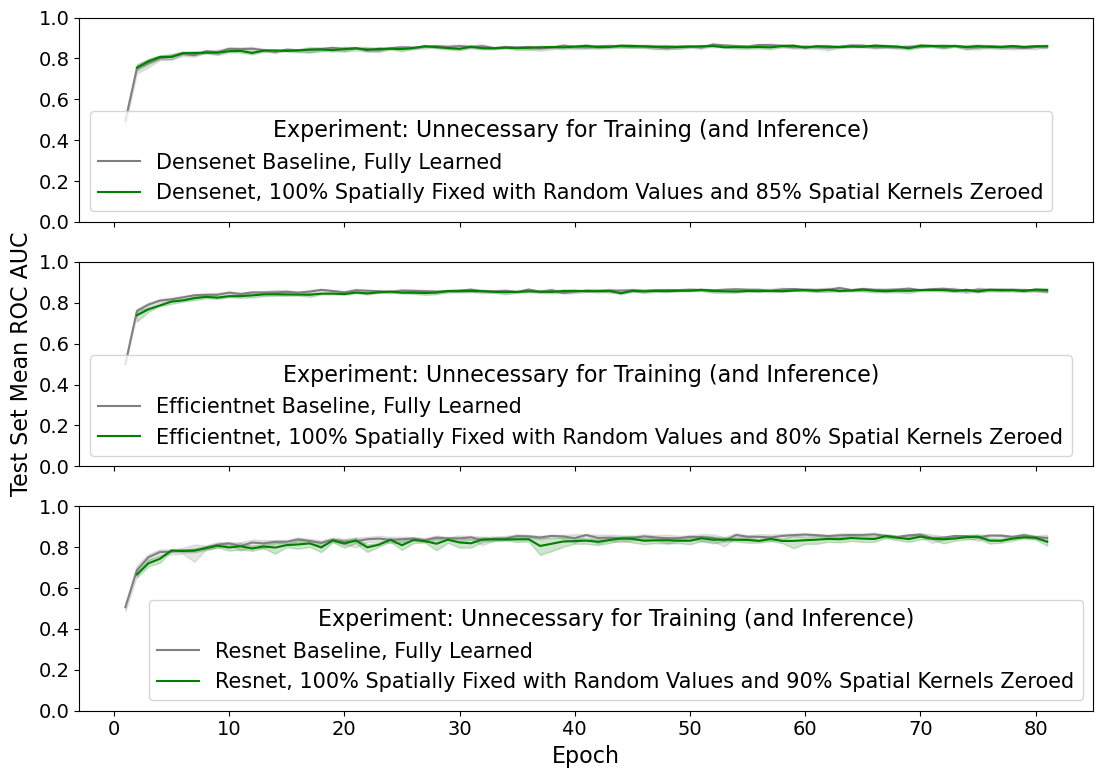}
  }
  \subfloat[Unnecessary for Inference\label{fig:unnecessary_inference}]{
    \includegraphics[width=.49\linewidth]{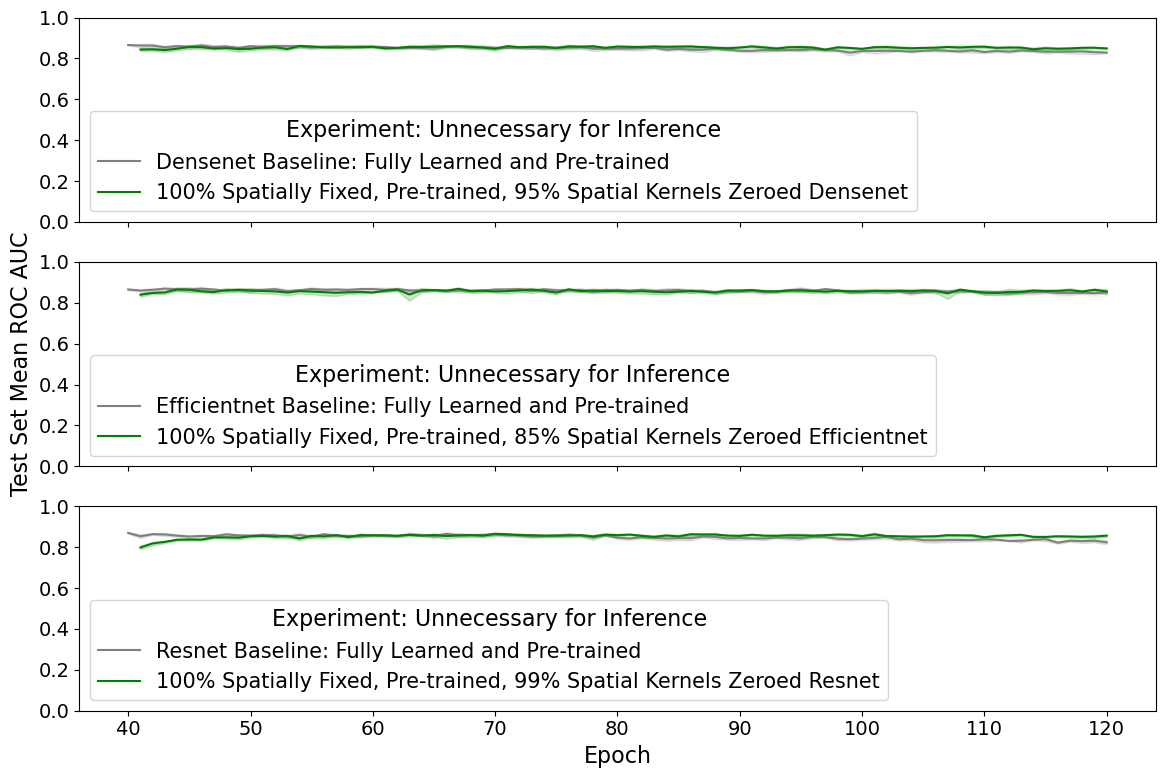}
  }
  \caption{\textbf{Spatial Filters are Unnecessary for Training and Inference}.  Nearly all spatial weights can be removed from the model by zeroing them out.  Spatial filters are unnecessary for both training \ref{fig:unnecessary_training} and inference \ref{fig:unnecessary_inference}.  Results hold for three architectures: DenseNet121 (top), EfficientNet-b0 (middle), ResNet50 (bottom).
    In all six subplots, each line represents an average with 95\% confidence interval of 6 independently trained models.  
    \\\\For Fig. \ref{fig:unnecessary_training}, we start with completely random deep networks with weights that have never been trained.  We fix the spatial weights and zero out 90\%, 85\% and 80\% of least salient spatial filter kernels for ResNet50, DenseNet121 and EfficientNet-b0 architectures, respectively.  We subsequently train the models on CheXpert.  The baselines are randomly initialized (never learned) deep network trained on CheXpert.  The results clearly show that only very few filters are necessary to the model.
    \\\\For Fig. \ref{fig:unnecessary_inference}, we start with one of our ImageNet Unchanged \ExplainFix models that was pre-trained on CheXpert.  We use the saliency method to zero out 99\%, 95\% and 85\% of least salient spatial filter kernels for ResNet50, DenseNet121 and EfficientNet-b0 architectures, respectively.   We then fine-tune the non-spatial weights on CheXpert.  The baselines are corresponding fully learned models that were pre-trained and fine-tuned on CheXpert.
  }
  \end{figure}

  \section{ChannelPrune: Sanity Check on Computational Savings}\label{appendix:comp_savings_same_acc}
\begin{figure}[H]
    \centering
    \subfloat[Faster to Train (same as Fig. \ref{fig:compute_efficiency2})\label{fig:compute_eff2}]{
      \includegraphics[width=.9\linewidth]{pics/zero_weights_timing_analysis.png}%
    }\\
    \subfloat[Nearly Equal Predictive Performance \label{fig:compute_eff2_acc}]{
      \includegraphics[width=.9\linewidth]{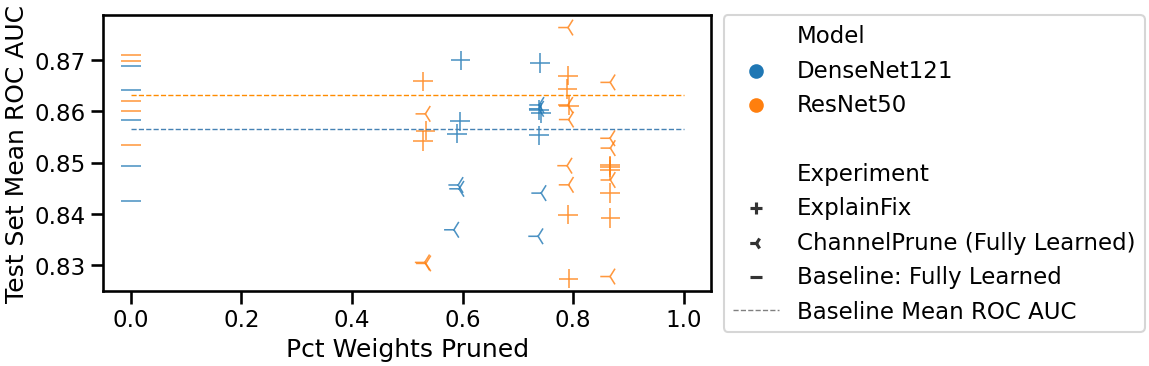}
    }
    \caption{\textbf{Nimble and Accurate.} \ExplainFix models are smaller, faster and as accurate.}
    \label{fig:compute_eff_vs_acc}
  \end{figure}

  As a sanity check that pruned and fixed models have predictive performance nearly equal to the unpruned baseline, we display the computational savings alongside a corresponding predictive performance plot Fig. \ref{fig:compute_eff_vs_acc}.  Fig. \ref{fig:compute_eff2} is identical to Fig. \ref{fig:compute_efficiency2}.  Each point in Fig. \ref{fig:compute_eff2_acc} is a separately trained model trained for 40 epochs and then evaluated on the CheXpert test set.  The horizontal dashed lines are the average of 5 independently trained fully learned and unpruned baselines.  All ChannelPrune and \ExplainFix models start from  a spatially fixed Unchanged ImageNet model that has never been trained on CheXpert.  The smallest ResNet50 model, at 86\% pruned parameters, is another sanity check that zeroing out too many spatial filters (99.9\% of them instead of 99\%) can cause modest performance loss.  In all other cases, the predictive performance of pruned fixed models is nearly equal to the unpruned baseline model, thus verifying the benefits of our proposed methods.

  \section{ChannelPrune: Percent Zeroed vs Percent Pruned} \label{appendix:table_pruning_zeroing}
  Given a starting spatial convolution layer, ChannelPrune removes input and output channels that are entirely zeroed.  A simple way to conceptualize the task is to consider the spatial filters as a tensor $(O,I,\dots)$, where $O$ and $I$ are the number of output and input channels.  If we reduce this to a boolean $O,I$ matrix indicating True if the spatial kernel is entirely zero, then ChannelPrune removes an input channel if if the entire column is True, and an output channel if the entire row is True.  Naturally, a zeroing process based on saliency weights will result in some rows that will no be pruned even though they are mostly zero.  We report the discrepancy between the percent of spatial filters zeroed and percent of all weights in the network that were pruned in the table below.  The analyzed models correspond to those in Appendix \ref{appendix:comp_savings_same_acc}.

\begin{table}[H]
  \centering
  \caption{\textbf{Illustration of ChannelPrune}.  The relationship between ChannelPrune and Saliency-based Zeroing.}
  \label{table:pruning_vs_zeroing}
  \begin{tabular}{l|c|c}
\toprule
Model & \% Spatially Zeroed  & \% Pruned\\
\midrule
DenseNet121 &       80.0 &            0.58 \\
DenseNet121 &       90.0 &            0.74 \\
ResNet50    &       90.0 &            0.52 \\
ResNet50    &       99.0 &            0.79 \\
ResNet50    &       99.9 &            0.86 \\
\bottomrule
  \end{tabular}
\end{table}

\printbibliography

\end{document}